\documentclass[10pt,twocolumn,letterpaper]{article}

\usepackage[pagenumbers]{cvpr}      
\usepackage{colortbl}
\usepackage{array}
\usepackage{tabularx}
\usepackage{makecell}
\newcommand{\fg}[1]{\mathbf{\mathcolor{ForestGreen}{#1}}}
\usepackage{graphicx}
\usepackage{multirow}
\usepackage{algorithm}
\usepackage{algorithmic}

\definecolor{cvprblue}{rgb}{0.21,0.49,0.74}
\usepackage[pagebackref,breaklinks,colorlinks,allcolors=cvprblue]{hyperref}

\def\paperID{3610} 
\def\confName{CVPR}
\def\confYear{2026}

\title{VLM-Pruner: Buffering for Spatial Sparsity in an Efficient VLM Centrifugal Token Pruning Paradigm}

\author{\textbf{Zhenkai Wu}$^{1,2}$\thanks{Equal contribution.} \quad
        \textbf{Xiaowen Ma}$^{2}$\footnotemark[1] \quad
        \textbf{Zhenliang Ni}$^{2}$\thanks{Corresponding authors: \{nizhenliang2; xinghao.chen\}@huawei.com} \\ 
        \textbf{Dengming Zhang}$^{1,2}$ \quad
        \textbf{Han Shu}$^{2}$ \quad
        \textbf{Xin Jiang}$^{2}$ \quad
        \textbf{Xinghao Chen}$^{2}$\footnotemark[2] \\
        $^1$ Zhejiang University \quad
        $^2$ Huawei Technologies
}

\begin{document}
\maketitle
\begin{abstract}
Vision–language models (VLMs) excel at image understanding tasks, but the large number of visual tokens imposes significant computational costs, hindering deployment on mobile devices. Many pruning methods rely solely on token importance and thus overlook inter-token redundancy, retaining numerous duplicated tokens and wasting capacity. Although some redundancy-aware approaches have been proposed, they often ignore the spatial relationships among visual tokens. This can lead to overly sparse selections of retained tokens that fail to adequately cover the regions of target objects.
To address these limitations, we propose VLM-Pruner, a training-free token pruning algorithm that explicitly balances redundancy and spatial sparsity. We introduce a centrifugal token pruning paradigm that enables near-to-far selection while prioritizing the preservation of fine-grained object details. Moreover, we design a Buffering for Spatial Sparsity (BSS) criterion that defers the selection of spatially distant tokens. We further adopt a parallel greedy strategy to conduct token selection efficiently. To mitigate information loss from pruning, we selectively fuse salient information from the discarded tokens into the retained ones.
Comprehensive comparisons demonstrate that VLM-Pruner consistently outperforms strong baselines across five VLMs with an 88.9\% pruning rate, while delivering an end-to-end inference speedup.
The code is available at \url{https://github.com/Casey-bit/VLMPruner}.
\end{abstract}

\section{Introduction}
\label{sec:intro}

\begin{figure}[h]
    \centering
    \includegraphics[width=\linewidth]{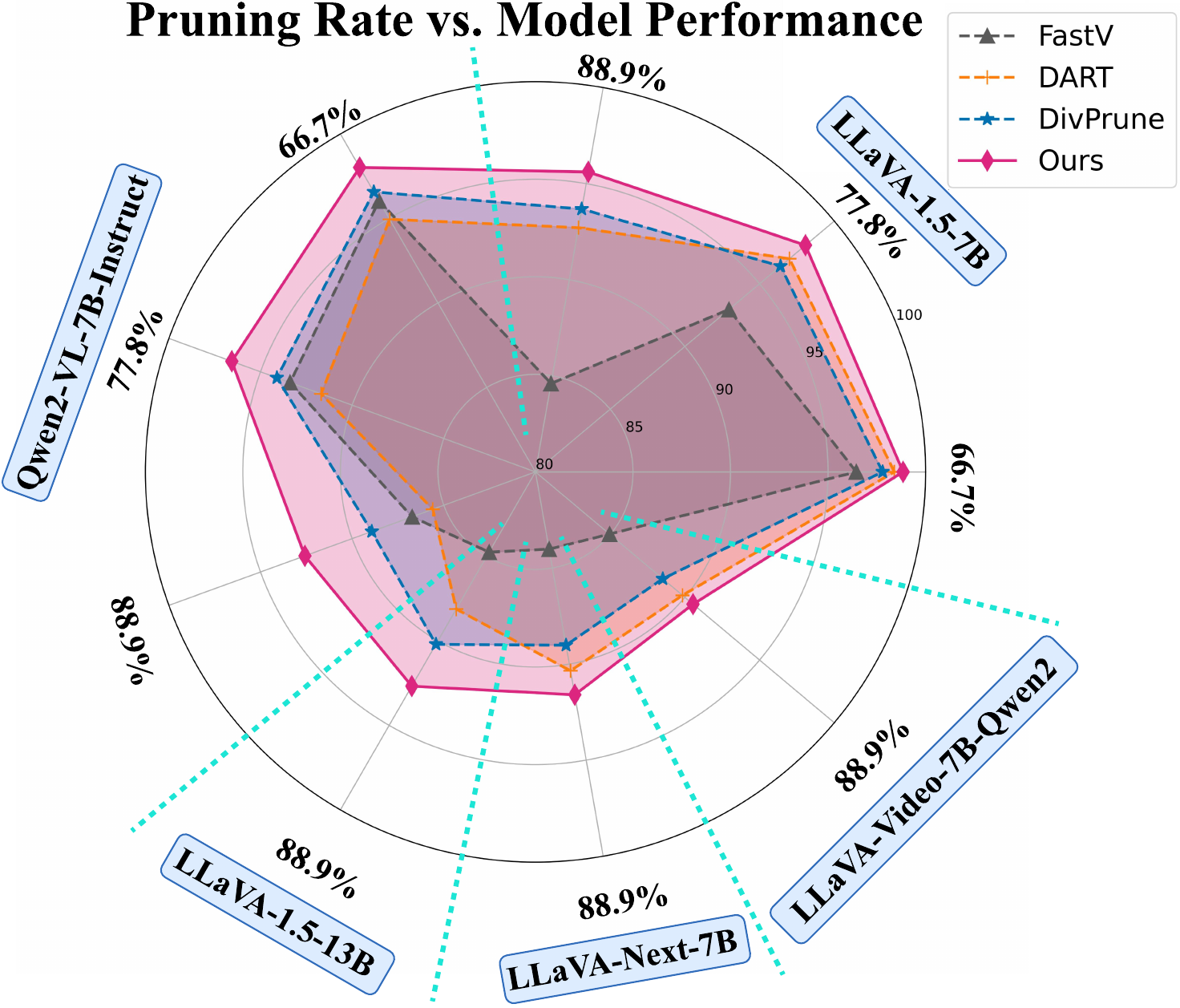}
    \caption{\textbf{Comparisons between baselines and VLM-Pruner.} Compared with importance-driven FastV and redundancy-reduction DART and DivPrune at pruning rates of 66.7\%, 77.8\%, and 88.9\%, VLM-Pruner consistently outperforms them across five VLMs.}
    \vspace{-3mm}
    \label{fig:intro1}
\end{figure}

\begin{figure}[h]
    \centering
    \includegraphics[width=\linewidth]{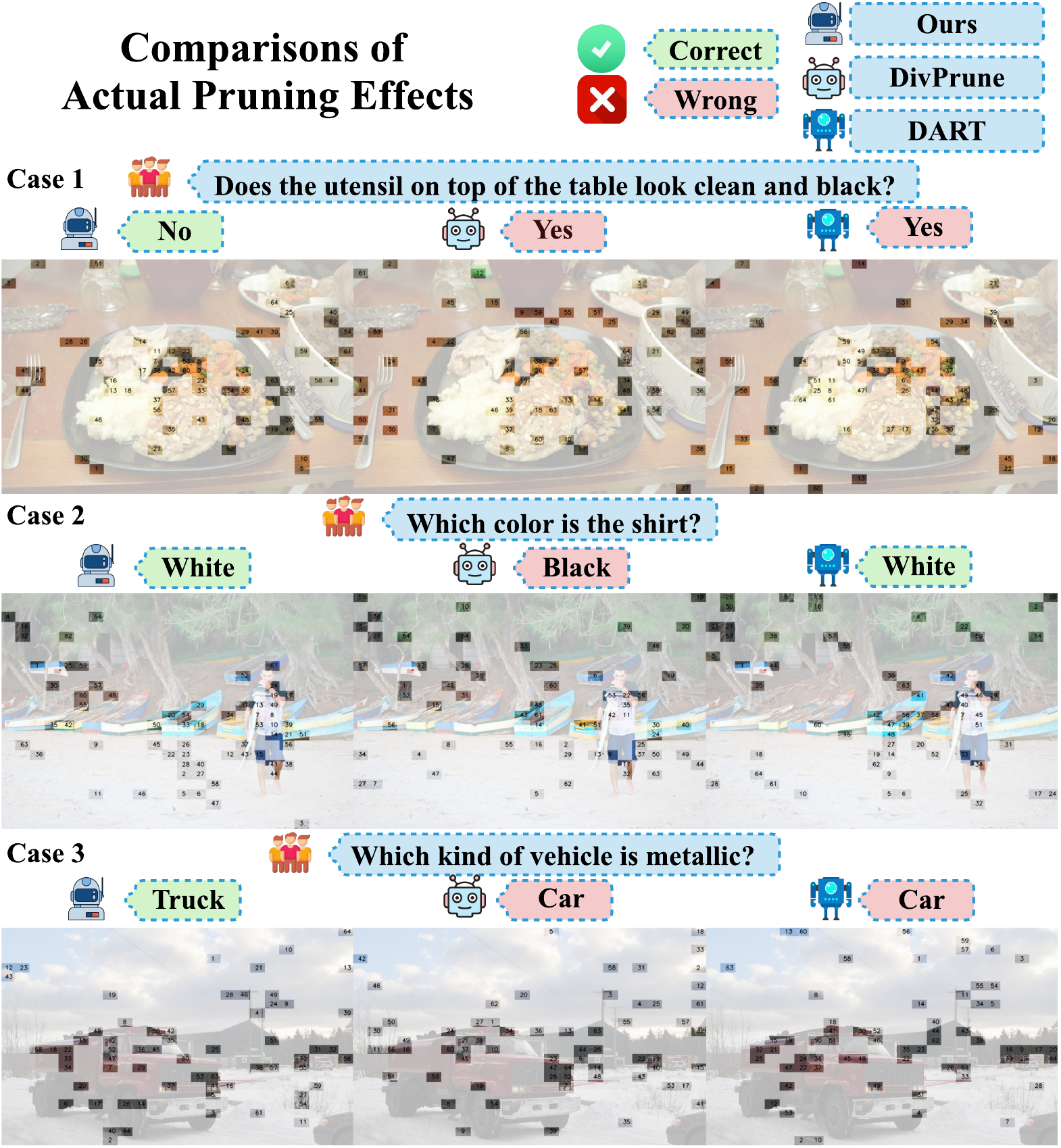}
    \caption{\textbf{Comparisons of the actual pruning effects between baselines and VLM-Pruner.} Visual question answering cases with correct (green) and incorrect (red) responses; numbers (from 1 to 64) denote selection order.}
    \vspace{-5mm}
    \label{fig:intro2}
\end{figure}

Building on the powerful reasoning capabilities of large language models (LLMs)~\cite{gpt4, vicuna, llama}, vision–language models (VLMs)~\cite{llava, llava15, llavanext, llavavideo, qwen2vl} integrate a visual encoder (e.g., ViT~\cite{vit}) to introduce visual modalities. This design enables VLMs to excel in tasks such as image captioning, visual question answering, video understanding, and multimodal reasoning. This integration significantly broadens their range of practical applications. However, the visual modality inherently carries high information density, particularly in high-resolution images \cite{llavanext} and multi-frame videos \cite{llavavideo}, which results in a number of visual tokens that can exceed textual tokens by hundreds or even thousands of times. When these abundant visual tokens are merely projected and concatenated with textual tokens before being fed into the decoder layers of LLMs \cite{llava}, where attention computation exhibits quadratic complexity, the overall efficiency is substantially degraded.

As discussed above, the excessive number of visual tokens not only reduces computational efficiency but also introduces significant redundancy: 1) Cross-modal redundancy, where only visual regions relevant to the textual query are necessary; and 2) Intra-modal redundancy, where tokens representing similar regions of the same object carry overlapping information. Recent studies tackle these issues with training-free token pruning that selects a representative subset of visual tokens, typically focusing on two strategies: 1) importance-driven pruning \cite{fastv, hired, pyramiddrop, sparsevlm,cdpruner}, which leverages intra- or cross-modal attention scores to retain salient tokens; and 2) redundancy-reduction pruning \cite{dart, divprune,btp,saint}, which employs greedy algorithms to iteratively choose tokens with low similarity to those selected.

However, the above strategies still perform suboptimally and exhibit notable drawbacks: \textbf{1) The importance-driven strategies focus on similar local regions often leads to redundancy.} Methods based on visual importance like FastV \cite{fastv} and on visual-language importance like SparseVLM \cite{sparsevlm} often perform no better than random token pruning. In some cases they even perform worse \cite{dart}, because importance driven criteria tend to preserve multiple similar local regions around the same focused object, causing substantial redundancy. Although methods such as FiCoCo~\cite{ficoco} and MustDrop~\cite{mustdrop} apply localized penalties to prevent redundancy from overly clustered tokens, as shown in the person of Case 2 in Fig.~\ref{fig:intro2}, moderate clustering can, in fact, better preserve local details.; \textbf{2) The redundancy-reduction strategies leads to dispersed token selections.} Methods targeting visual redundancy like DivPrune~\cite{divprune} and those addressing cross-modal redundancy like DART~\cite{dart} generate token selections that are dispersed yet incomplete, failing to capture the fine-grained details of objects. As shown in Fig. \ref{fig:intro2}, edge background regions have low similarity to the main object, so DivPrune and DART are prone to selecting edge background tokens and jumping erratically between background and foreground, resulting in a scattered selection.

Accordingly, we propose a training-free centrifugal token pruning paradigm, termed VLM-Pruner, that effectively balances token redundancy and local-detail completeness. VLM-Pruner follows a near-to-far token selection process, starting with pivot tokens, then prioritizing spatially adjacent tokens, and finally recovering from discarded tokens. This ensures an orderly sequence and avoids the chaotic, incomplete coverage from arbitrary selection. For example, in the first case of Fig. \ref{fig:intro2}, four tokens closely aligned in order on the fork at the left of the image are selected when compared to other baselines; in the third case, VLM-Pruner captures comprehensive truck details, including the body, tires, and windshield.  Specifically, a minimal pivot token set with maximum minimum distance \cite{maxmin} is firstly identified to coarsely represent distinct subjects. Secondly, the Buffering for Spatial Sparsity (BSS) criterion leverages the minimum spatial distance between candidate tokens and the selected set to preferentially select spatially adjacent tokens with low redundancy, thereby ensuring an orderly selection process. Tokens are then greedily chosen in ascending similarity order, with parallel processing for acceleration. Finally, the remaining tokens are matched to the selected set based on maximum similarity and merged through Similarity-Weighted Aggregation (SWA) to recover the outermost information. In summary, our major contributions are as follows:

\begin{itemize}
    \item We propose VLM-Pruner, a training free centrifugal token pruning paradigm for efficient VLMs that balances token redundancy and local-detail completeness.
    \item We adopt the Buffering for Spatial Sparsity criterion to enforce relatively ordered token pruning and effectively alleviate scattered token distribution.
    \item We conduct extensive experiments on 13 benchmarks across 5 VLMs. As shown in Fig. \ref{fig:intro1}, VLM-Pruner delivers state-of-the-art results, with gains that grow as the pruning ratio increases.
\end{itemize}

\begin{figure*}[t]
    \vspace{-2mm}
    \centering
    \includegraphics[width=\linewidth]{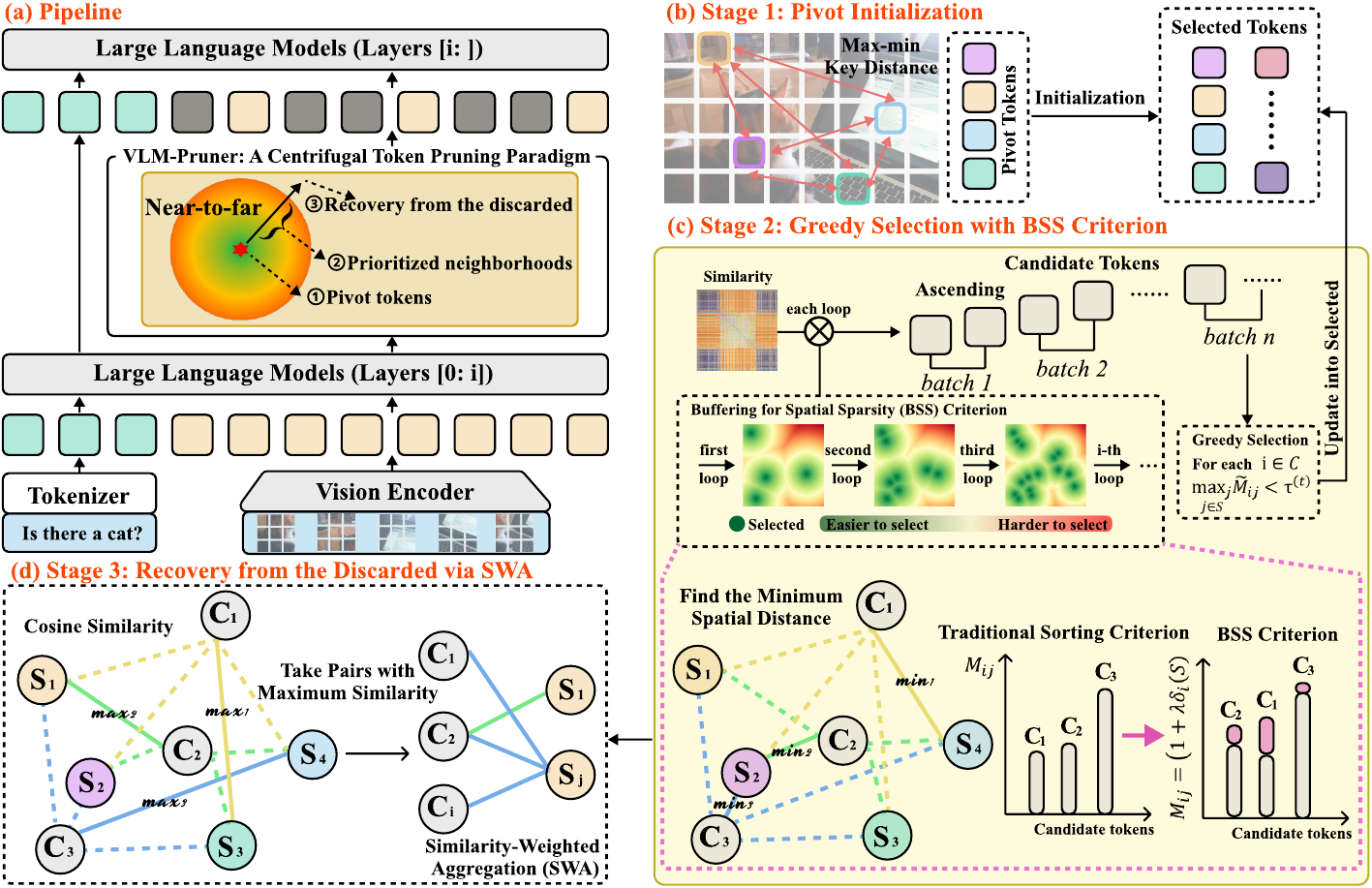}
    \caption{\textbf{Centrifugal token pruning paradigm of VLM-Pruner.} \textbf{(a) Pipeline:} In the $i$-th decoder layer of the LLM, VLM-Pruner follows a near-to-far selection order, \textbf{(b)} starting with pivot tokens, \textbf{(c)} gradually expanding outward from neighborhoods, and \textbf{(d)} ultimately recovering the outermost information from the discarded tokens via SWA. The similarity computed under BSS criterion makes candidate tokens spatially closer to selected ones more likely to be chosen. Color transition from green to red indicates decreasing selection probability. $C$ and $S$ denote candidate and selected tokens, respectively. After applying BSS, the closer candidate $C_2$ is prioritized over $C_1$.
}
    \vspace{-3mm}
    \label{fig:centrifugal}
\end{figure*}

\section{Related Work}
\label{sec:related}

\subsection{Efficient Vision-Language Models}
VLMs \cite{vit} align visual encoders with LLMs for captioning, VQA, grounding, and reasoning, spanning image-based \cite{llava, llava15, llavanext, qwen2vl} and video-based \cite{llavavideo} variants. A common approach pairs an LLM with a high-capacity visual encoder such as ViT \cite{vit}, CLIP-ViT \cite{clipvit}, or SigLIP \cite{siglip}, and connects them through cross-modal alignment.

Efficiency efforts aim to preserve multimodal capability while cutting compute, memory, and latency. (i) Connector-tuning freezes the backbones and trains lightweight adapters, e.g., BLIP-2 \cite{blip}, avoiding full end-to-end retraining. (ii) Content-adaptive budgeting allocates resolution and token budgets based on input saliency, as in Qwen2-VL \cite{qwen2vl}. (iii) Distill-and-prune compression transfers knowledge \cite{kd} and removes modality-aware redundancy. EfficientVLM \cite{efficientvlm} retains 44.3\% parameters and yields the inference speedup with minimal quality loss. Expert Merging \cite{zhang2025expert} integrates capabilities of multiple VLMs.

\subsection{Token Pruning}
Token pruning reduces quadratic attention cost by compressing tokens at the source. ToMe \cite{tome} introduced progressive token merging for ViTs \cite{vit}, motivating VLM pruning methods that are training-required or training-free. ATP-LLaVA \cite{atpllava} learns layer-wise, instance-specific pruning ratios with MLPs but adds training overhead. We instead focus on training-free pruning, which mainly includes importance-driven and redundancy-reduction methods.

Importance-driven methods retain high-attention tokens. FastV \cite{fastv} uses visual self-attention, SparseVLM \cite{sparsevlm} prunes visual and text tokens via cross-modal attention, and HiRED \cite{hired} and MustDrop \cite{mustdrop} keep tokens that strongly attend to [CLS]. PDrop \cite{pyramiddrop} selects tokens attending to the last instruction token, while CDPruner \cite{cdpruner} ranks tokens by joint relevance to both text and visual tokens.
Redundancy-reduction methods emphasize diversity. DART \cite{dart} incrementally selects tokens with low similarity to text and visual anchors, while DivPrune \cite{divprune} and BTP \cite{btp} greedily keep tokens that are least similar to those already selected. SAINT \cite{saint} likewise favors dissimilar tokens.

In practice, importance-driven pruning can overselect semantically overlapping tokens, whereas redundancy-reduction selection may be unstable when similarity gaps between background and salient regions are large, yielding dispersed coverage and missing fine-grained details.

\section{Methodology}
\label{sec:method}


\subsection{Overview}
As shown in Fig.~\ref{fig:centrifugal}, we present a training-free centrifugal pipeline for retaining the most informative image tokens in VLMs, while ensuring comprehensive coverage of dense local object details. 
Specifically, VLM-Pruner follows a three-stage paradigm:
(i) initialize a small set of diverse pivots in feature space via a max-min rule;
(ii) expand the retained set with a parallel greedy procedure regularized by a BSS criterion that prioritizes densifying local neighborhoods before expanding outward; and
(iii) re-inject the information of discarded tokens into their most similar retained pivots through a similarity-weighted aggregation.

\subsection{Buffering for Spatial Sparsity (BSS) Criterion}
We perform VLM-Pruner on the second layer of the LLM decoder to enforce an ordered near-to-far selection process, thereby preventing the token distribution from becoming too scattered. Given that the visual feature map has a spatial size of $H \times W$, it results in $N = H \times W$ tokens. We denote the token $keys$ as $\mathbf{K} \in \mathbb{R}^{N \times d_k}$ to define the identity of features, and the $hidden\_states$ as $\mathbf{H} \in \mathbb{R}^{N \times d}$ to compute similarities, respectively.
The spatial coordinate of token $i\!\in\!\{0,\dots,N - 1\}$ is $\mathbf{p}_i=(x_i,y_i)$ with
\begin{equation}
x_i \;=\; i \bmod W, \quad y_i \;=\; \bigl\lfloor i / W \bigr\rfloor .
\end{equation}

The grid distance and its maximum can be defined as
\begin{equation}
    D^{(\mathrm{sp})}_{ij} \;=\; \bigl\lVert \mathbf{p}_i-\mathbf{p}_j \bigr\rVert_2,
    \quad
    D_{\max} \;=\; \sqrt{H^2+W^2}.
\end{equation}

To reduce the cost of a full $N \times N$ similarity, we first screen the channels based on variance. We compute the per-channel variance of $\mathbf{H}$ across tokens, then keep the top-$q$ channels (with $q = 256$). Afterward, we form the reduced features as 
\(
\tilde{\mathbf{H}} \,=\, \mathbf{H}_{:\,\!,\mathcal{I}} \in \mathbb{R}^{N\times q}.
\)
We use cosine similarity
\begin{equation}
    M_{ij} \;=\;
    \frac{\tilde{\mathbf{H}}_i^\top \tilde{\mathbf{H}}_j}
         {\lVert \tilde{\mathbf{H}}_i\rVert_2\,\lVert \tilde{\mathbf{H}}_j\rVert_2},
    \label{eq:cos}
\end{equation}
and precompute the matrix $M\in\mathbb{R}^{N\times N}$ once.

Our central claim is that prioritizing spatial proximity during token selection facilitates a more complete reconstruction of fine-grained object details. 
In other words, the selection process naturally follows a centrifugal paradigm, starting from locality and gradually expanding outward.

We denote $\mathcal{V} = \{0, \dots, N - 1\}$ as the set of all tokens. We denote $\mathcal{S} \subseteq \mathcal{V}$ as the retained set, where $|\mathcal{S}| = \mathcal{R}$. We also denote $\mathcal{C} = \mathcal{V} \setminus \mathcal{S}$ as the candidate set, where $|\mathcal{C}| = N - \mathcal{R}$.
We pose selection as a maximum diversity problem $\mathcal{F}$  with a constraint of spatial sparsity:
\begin{equation}
\max_{S\subseteq V}\;\; 
\mathcal{F}(S) 
\;= \;
\max\!\!\sum_{i\subseteq \mathcal{C},j\subseteq S}\! \min\widetilde{M}_{ij},
\label{eq:pbfl}
\end{equation}
where $\widetilde{M}_{ij}$ is a BSS-modulated similarity
\begin{align}
\delta_i(S) \;&=\; \min_{j\in S} D^{(\mathrm{sp})}_{ij},
\nonumber
\\
\bar{\delta}_i(S) \;&=\; \delta_i(S) / D_{\max},
\label{eq:bss}
\\
\widetilde{M}_{ij} \;&=\; M_{ij}\,\bigl(1+\lambda\,\bar{\delta}_i(S)\bigr).
\nonumber
\end{align}

When $\lambda = 0$, the objective $\mathcal{F}$ reduces to a standard monotone submodular function \cite{cost}, for which the greedy algorithm achieves the classical approximation guarantee \cite{greedy}. 
When $\lambda > 0$, as the iterations proceed, the number of tokens in the selected set $\mathcal{S}$ gradually increases, causing the remaining candidate tokens to lie closer to the already selected ones in spatial space. 
Consequently, the normalized spatial distance term $\bar{\delta}_i(S)$ in Eq.~\eqref{eq:bss} continuously decreases with each iteration. 
This monotonic decay preserves the overall submodular structure of $\mathcal{F}$ across iterations and therefore still permits an efficient greedy optimization scheme.
Meanwhile, for $\lambda > 0$, tokens that are spatially farther from the already selected set are assigned a larger scaling coefficient in their similarity computation, making them less likely to be selected in subsequent iterations. 
As shown in Fig. \ref{fig:centrifugal}, this adaptive weighting mechanism effectively realizes the BSS effect, ensuring that nearby tokens are prioritized first, thereby promoting dense local coverage before the selection expands outward.

\begin{algorithm}[t]
\caption{Centrifugal Token Pruning Paradigm with BSS}
\label{alg:ours}
\small
\begin{algorithmic}[1]
\REQUIRE $\mathbf{K}\!\in\!\mathbb{R}^{N\times d_k}$, $\mathbf{H}\!\in\!\mathbb{R}^{N\times d}$, coordinates $\{\mathbf{p}_i\}$; target token count $R$, pivot count $\kappa$, kept channels $q$, BSS strength $\lambda$, threshold schedule $\tau^{(0)}$, $\Delta\tau$, batch size $B$.
\STATE Keep the top-$q$ channels with the highest variance to form $\tilde{\mathbf{H}}$; build the similarity matrix $M$ by Eq. \eqref{eq:cos}.
\STATE Max--min initialize $\kappa$ pivots to get $\mathcal{S}_\kappa$ via \eqref{pivot}.
\STATE Precompute $D^{(\mathrm{sp})}_{ij}$ and $D_{\max}$.
\STATE $t\!\leftarrow\!0$, $\tau\!\leftarrow\!\tau^{(0)}$, $\mathcal{S}\!\leftarrow\!\mathcal{S}_\kappa$, $j \in \mathcal{S}$.
\WHILE{$|\mathcal{S}| < R$}
    \STATE $\mathcal{C}\!\leftarrow\!\{1,\dots,N\}\setminus\mathcal{S}$.
    \STATE For each $i\!\in\!\mathcal{C}$, compute $\bar{\delta}_i$ and form $\widetilde{M}_{ij}$ via Eq. \eqref{eq:bss}.
    \STATE Compute $r_i$ by Eq. \eqref{eq:nondup}; sort $i$ in descending $r_i$.
    \FOR{batches of $B$ candidates $i$}
        \item Recompute $\bar{\delta}_i$ and $\widetilde{M}_{ij}$ via Eq. \eqref{eq:bss}.
        \IF{$\max_{j\in\mathcal{S}} \widetilde{M}_{ij} < \tau$}
            \STATE $\mathcal{S}\leftarrow \mathcal{S}\cup\{i\}$; \textbf{break} if $|\mathcal{S}|=R$.
        \ENDIF
    \ENDFOR
    \STATE \textbf{break} if no token added this loop.
    \STATE $t\!\leftarrow\!t+1$, $\tau\!\leftarrow\!\tau+\Delta\tau$.
\ENDWHILE
\STATE $\mathcal{D}\!\leftarrow\!\{1,\dots,N\}\setminus\mathcal{S}$; build clusters $\{\mathcal{D}_j\}$ via $j^\star(u)$.
\STATE Compute $\mathbf{M}_j$ and update $\mathbf{H}_j$ by Eq. \eqref{eq:agg}.
\RETURN Selected indices $\mathcal{S}$ and updated $\mathbf{H}$.
\end{algorithmic}
\end{algorithm}

\subsection{Overall Token Pruning Pipeline}
We briefly summarize our proposed centrifugal token pruning method, termed VLM-Pruner, in Algorithm \ref{alg:ours}.

\subsubsection{Stage 1: Pivot Initialization}
\label{sec:stage1}
This initialization yields widely separated pivots that coarsely cover different semantic regions.

We denote $\kappa$ as the number of initial pivots. The token $keys$ inherently summarize the semantic characteristics of each token \cite{attention} and contain less redundant information due to their lower dimensionality. 
Therefore, we select $\kappa$ diverse pivots $\mathcal{S}_\kappa$ in the token $keys$ space using a max-min strategy:
\begin{align}
    j_1 &= \arg\max_{j\in V} \lVert \mathbf{K}_j \rVert_1, 
    \nonumber
    \\
    j_t &= \arg\max_{j \in \mathcal{C}} \min_{j'\in\mathcal{S}_{t-1}}
            \bigl\lVert \mathbf{K}_j - \mathbf{K}_{j'} \bigr\rVert_2,
            \quad t=2,\dots,\kappa, 
    \label{pivot}        
    \\
    \mathcal{S}_t &\leftarrow \mathcal{S}_{t-1}\cup\{j_t\}.
    \nonumber
\end{align}

Finally, we define the selected set as $\mathcal{S} \leftarrow \mathcal{S}_\kappa$.

\subsubsection{Stage 2: Greedy Selection with BSS Criterion}
\label{sec:stage2}
A purely redundancy-reduction greedy selection may over-disperse tokens. 
To address this, we introduce the BSS criterion, which enforces an ordered near-to-far selection process that prioritizes spatially proximate tokens.

We therefore modulate candidate-to-selected similarities by spatial sparsity.
For each candidate $i\notin\mathcal{S}$, compute the normalized nearest-selected distance $\bar{\delta}_i$,
and apply the BSS gain in Eq. \eqref{eq:bss} with a $\lambda > 0$ (we use $\lambda=0.5$).
Because $\bar{\delta}_i$ grows with spatial distance from $\mathcal{S}$, \eqref{eq:bss} makes farther candidates appear more similar to some selected token. So they are judged more redundant and thus deferred.
This yields a near-to-far centrifugal growth that first fills local details then expands outward.

Given $i \in\mathcal{C}$, the non-duplication score is defined as:
\begin{equation}
    r_i \;=\; 1 - \max_{j\in\mathcal{S}} \widetilde{M}_{ij},
    \label{eq:nondup}
\end{equation}
We then sort the candidates by $r_i$ in descending order and process them in parallel batches of size $B$, where $\widetilde{M}_{ij}$ is recomputed within each batch.
A candidate $i$ will be accepted in loop $t$ if:
\begin{equation}
    \max_{j\in\mathcal{S}} \widetilde{M}_{ij} \;<\; \tau^{(t)},
\label{eq:thres}
\end{equation}
where,
\(
     \tau^{(0)}=0.8,\; \tau^{(t{+}1)}=\tau^{(t)}+\Delta\tau,\; \Delta\tau=0.1.
\)

Due to the BSS gain, the upper bound of $\widetilde{M}_{ij}$ increases from $1$ to $(1+\lambda)$.
The strict-to-loose schedule prevents early admission of isolated, far-away tokens.
We iterate loops until $|\mathcal{S}|=R$ or no new tokens are added.

\subsubsection{Stage 3: Recovery from the Discarded via SWA}
\label{sec:stage3}

While the centrifugal selection process effectively preserves more complete details, it inevitably discards many outermost tokens that may still contain complementary semantic cues. 
To mitigate this loss, we introduce a Similarity-Weighted Aggregation (SWA) mechanism that recovers useful information from discarded tokens and reintegrates it into the most relevant retained tokens.

We denote the discarded set as $\mathcal{D}=\{0,\dots,N - 1\}\setminus\mathcal{S}$. 
Each discarded token $u\in\mathcal{D}$ is assigned to its most similar retained token:
\begin{equation}
    j^\star(u)=\arg\max_{j\in\mathcal{S}} M_{uj},
\end{equation}
which forms clusters $\mathcal{D}_j=\{u\in\mathcal{D} \,\And\, j^\star(u)=j\}$ for each retained token $j\in\mathcal{S}$.
We then compute similarity-normalized weights for each cluster $\mathcal{D}_j$ and aggregate the hidden states $\mathbf{H}_u$ of the discarded tokens:
\begin{align}
    \alpha_{u\to j}
        &= \frac{M_{uj}}{\sum_{u'\in\mathcal{D}_j} M_{u'j} + \varepsilon}, 
        \nonumber \\[2pt]
    \mathbf{E}_j
        &= \sum_{u\in\mathcal{D}_j} \alpha_{u\to j}\, \mathbf{H}_u, 
        \label{eq:agg} \\
    \mathbf{H}_j
        &= \beta\,\mathbf{H}_j + (1-\beta)\,\mathbf{E}_j,
        \qquad \beta\in(0,1),
        \nonumber
\end{align}
where $\alpha_{u\to j}$ denotes the normalized aggregation weight between a discarded token $u$ and its assigned retained token $j$, and $\mathbf{E}_j$ represents the similarity-weighted hidden state aggregated from all discarded tokens mapped to $j$. 
The term $\varepsilon = 10^{-8}$ ensures numerical stability. 
We set $\beta = 0.3$ in practice, meaning that $70\%$ of the aggregated hidden state $\mathbf{E}_j$ is incorporated into the final representation of the retained token.

\subsubsection{Complexity}
Channel screening and similarity construction cost $O(N d)$ and $O(N^2 q)$, respectively. Spatial distances cost $O(N^2)$ once.
Each greedy loop performs batched tensor ops equivalent to $O(N|\mathcal{S}|)$ (indexing and selecting the minimum spatial distance). The number of loops is small in practice due to the annealed threshold.
The aggregation stage performs a most similar assignment (using \texttt{argmax}) and a similarity-weighted matrix-vector update for each retained token.

\definecolor{mygray}{gray}{.92}
\definecolor{ForestGreen}{RGB}{34,139,34}
\definecolor{Forestred}{RGB}{220,50,50}
\begin{table*}[t]
    \centering
    \setlength{\tabcolsep}{5pt}
    \renewcommand{\arraystretch}{0.9}
    \footnotesize
    \centering
    \begin{tabular}{c | c c c c c c c c c| >{\centering\arraybackslash}p{1.0cm}}
        \toprule[1.5pt]
        \textbf{Method} & \makecell{\textbf{GQA}\\\scriptsize{Acc.}}& \makecell{\textbf{MMB}\\\scriptsize{Acc.}} & \makecell{\textbf{MME}\\\scriptsize{P+C-score}} & \makecell{\textbf{POPE}\\\scriptsize{F1}} & \makecell{{\textbf{SQA}}\\\scriptsize{IMG-Acc.}} &  \makecell{{\textbf{VQA}$^{\text{Text}}$}\\\scriptsize{Acc.}} & \makecell{\textbf{OCRBench}\\\scriptsize{Final-score}} & \makecell{{\textbf{SEED}$^{\text{Image}}$}\\\scriptsize{Acc.}} & \makecell{\textbf{OK-VQA}\\\scriptsize{VQA-score}}& \makecell[c]{\textbf{Avg}.}\\
        \hline
        \rowcolor{mygray}
          \multicolumn{11}{c}{\textit{Upper Bound, 576  Tokens} \ $\textbf{(100\%)}$}\\
         \textcolor{gray}{LLaVA-1.5-7B} & \textcolor{gray}{61.95}  & \textcolor{gray}{64.7} &  \textcolor{gray}{1862} & \textcolor{gray}{85.9} & \textcolor{gray}{69.46} & \textcolor{gray}{58.17} & \textcolor{gray}{297} &
         \textcolor{gray}{66.18} &
         \textcolor{gray}{57.98} &
         \textcolor{gray}{100.0\%} \\
          \hline
       \rowcolor{mygray}
         \multicolumn{11}{c}{\textit{Retain 192 Tokens} \ $\fg{(\downarrow 66.7\%)}$} \\

        FastV \scriptsize{(ECCV24)} & 57.94 & 61.2 & \underline{1829} & 78.2 & 68.77 & 57.37 & 293 & 63.60 & 56.97 & 96.45\% \\
        SparseVLM \scriptsize{(ICML25)} & 59.50 & 62.5 & 1782 & 85.7 & 68.57 & 57.70 & 294 & 64.23 & \underline{57.57} & 97.93\%  \\
        PDrop \scriptsize{(CVPR25)} & 57.19 & 63.2 & 1787 & 80.3 & 69.11 & 56.16 & 278 & 59.34 & 56.01 & 95.04\% \\
        MustDrop \scriptsize{(arXiv24)}  & 58.20 & 62.3 & 1728 & 83.6 & \underline{69.20} & 55.87 & \bf314 & 62.77 & 56.57 & 97.13\%\\
        DART \scriptsize{(EMNLP25)} & 60.08 & \bf63.6 & \bf1855 & 82.9 & 69.06 & \underline{57.52} & \underline{296} & \underline{64.62} & 57.17 & \underline{98.40\%}\\
        DivPrune \scriptsize{(CVPR25)} & 60.02 & 62.0 & 1771 & \bf87.7 & 68.72 & 56.64 & 293 & 63.89 & 57.29 & 97.80\% \\

        \rowcolor{orange!20}
        Ours & \bf61.31 & \underline{63.5} & 1788 & \underline{86.0} & \bf69.96 & \bf57.97 & 290 & \bf65.34 & \bf57.78 & \bf98.85\% \\

        \hline
       \rowcolor{mygray}
         \multicolumn{11}{c}{\textit{Retain 128 Tokens} \ $\fg{(\downarrow 77.8\%)}$} \\

        FastV \scriptsize{(ECCV24)} & 55.86 & 56.1 & 1721 & 72.2 & 68.86 & 56.95 & 288 & 61.41 & 55.87 &  92.95\% \\
        SparseVLM \scriptsize{(ICML25)} & 58.42 & 60.0 & 1753 & \underline{85.5} & 68.67 & 56.67 & 278 & \underline{63.63} & \underline{56.78} & 96.08\%  \\
        PDrop \scriptsize{(CVPR25)} & 54.43 & 61.1 & 1611 & 73.1 & 69.05 & 53.59 & 268  & 56.44 & 53.41 & 90.34\%\\
        MustDrop \scriptsize{(arXiv24)} & 56.19 & 61.1 & 1690 & 79.1 & 68.50 & 56.04 & \bf306 & 60.53 & 55.48 & 94.80\%\\
        DART \scriptsize{(EMNLP25)} & 58.70 & \bf63.1 & \bf1826 & 80.7 & 68.96 & 56.42 & \underline{294} & 63.41 & 56.58 & \underline{97.00\%}\\
        DivPrune \scriptsize{(CVPR25)} & \underline{59.35} & 61.6 & 1713 & \bf87.5 & 68.67 & 55.92 & 285 & 62.34 & 56.56 & 96.40\%\\

        \rowcolor{orange!20}
        Ours & \bf61.00 & \underline{62.5} & \underline{1768} & 85.4 & \bf69.31 & \bf57.32 & \underline{294} & \bf64.35 & \bf57.18 & \bf98.07\%\\

        \hline
       \rowcolor{mygray}
           \multicolumn{11}{c}{\textit{Retain 64 Tokens} \ $\fg{(\downarrow 88.9\%)}$} \\

        FastV \scriptsize{(ECCV24)} & 51.66 & 48.0 & 1564 & 59.7 & 67.91 & \underline{55.35} & 249 & 55.48 & 52.04 & 84.60\%  \\
        SparseVLM \scriptsize{(ICML25)} & 53.75 & 56.2 & 1593 & 78.0 & \bf69.71 & 53.47 & 203 & 56.75 & 53.42 & 87.61\%  \\
        PDrop \scriptsize{(CVPR25)} & 41.9 & 33.3 & 1092 & 55.9 & 68.60 & 45.90 & 250 & 54.73 & 52.98 & 75.42\%\\
        MustDrop \scriptsize{(arXiv24)}  & 52.71 & 60.0 & 1533 & 68.6 & \underline{68.96} & 54.49 & \underline{278} & 54.68 & 53.48 & 89.05\%\\
        DART \scriptsize{(EMNLP25)}  & 56.38 & \underline{60.6} & \underline{1751} & 74.6 & 68.67 & 54.30 & 271 & 59.29 & 55.52 & 92.71\% \\
        DivPrune \scriptsize{(CVPR25)}& \underline{57.74} & 60.1 & 1619 & \bf86.2 & 67.97 & 54.60 & 271 & \underline{60.19} & \underline{55.57} & \underline{93.68\%}  \\

        \rowcolor{orange!20}
        Ours & \bf59.15 & \bf61.3 & \bf1752 & \underline{82.2} & 68.37 & \bf56.05 & \bf279 & \bf62.24 & \bf56.63 & \bf95.61\% \\

        \bottomrule[1.5pt]
	\end{tabular}
	\caption{Comparative experiments on image understanding are performed on LLaVA-1.5-7B.}
    \label{tab:1}
\end{table*}

\section{Experiment}

\definecolor{mygray}{gray}{.92}
\definecolor{ForestGreen}{RGB}{34,139,34}
\definecolor{Forestred}{RGB}{220,50,50}
\begin{table*}[t]
    \centering
    \setlength{\tabcolsep}{5pt}
    \renewcommand{\arraystretch}{0.9}
    \footnotesize
    \centering
    \begin{tabular}{c | c c c c c c c c c| >{\centering\arraybackslash}p{1.0cm}}
        \toprule[1.5pt]
        \textbf{Method} & \makecell{\textbf{GQA}\\\scriptsize{Acc.}}& \makecell{\textbf{MMB}\\\scriptsize{Acc.}} & \makecell{\textbf{MME}\\\scriptsize{P+C-score}} & \makecell{\textbf{POPE}\\\scriptsize{F1}} & \makecell{{\textbf{SQA}}\\\scriptsize{IMG-Acc.}} &  \makecell{{\textbf{VQA}$^{\text{Text}}$}\\\scriptsize{Acc.}} & \makecell{\textbf{OCRBench}\\\scriptsize{Final-score}} & \makecell{{\textbf{SEED}$^{\text{Image}}$}\\\scriptsize{Acc.}} & \makecell{\textbf{OK-VQA}\\\scriptsize{VQA-score}}& \makecell[c]{\textbf{Avg}.}\\
        \hline
        \rowcolor{mygray}
          \multicolumn{11}{c}{\textit{Upper Bound, 576  Tokens} \ $\textbf{(100\%)}$}\\
         \textcolor{gray}{LLaVA-1.5-13B} & \textcolor{gray}{63.22}  & \textcolor{gray}{68.6} &  \textcolor{gray}{2826} & \textcolor{gray}{86.4} & \textcolor{gray}{72.68} & \textcolor{gray}{61.19} & \textcolor{gray}{331} &
         \textcolor{gray}{68.23} &
         \textcolor{gray}{60.87} &
         \textcolor{gray}{100.0\%} \\
          \hline
       \rowcolor{mygray}
         \multicolumn{11}{c}{\textit{Retain 64 Tokens} \ $\fg{(\downarrow 88.9\%)}$} \\

        FastV \scriptsize{(ECCV24)} & 55.75 & 55.1 & 1670 & 69.5 & \underline{72.58} & 56.1 & 270 &59.81 & 57.19 & 84.75\%   \\
        DART \scriptsize{(EMNLP25)} &57.06  & \bf65.7 &1736 &76.0 & \bf73.87 &55.84 &264 &61.76 &\underline{57.54} &88.12\%\\
        DivPrune \scriptsize{(CVPR25)} & \underline{57.93} & \underline{64.3} & \underline{1754} & \bf84.8 &71.64 &\underline{57.38} &\underline{287} &\underline{62.31} &57.32 &\underline{90.20\%} \\

        \rowcolor{orange!20}
        Ours & \bf61.21 & \bf65.7 & \bf1779 & \underline{84.7} & 72.43 & \bf58.74 & \bf306 & \bf64.91 &\bf59.23 & \bf92.68\% \\
        \hline
        \rowcolor{mygray}
          \multicolumn{11}{c}{\textit{Dynamic Resolution (Max pixels = 1344 $\times$ 336), Dynamic Height and Width}}\\
          \rowcolor{mygray}
          \multicolumn{11}{c}{\textit{Upper Bound, Dynamic Tokens, Max Tokens = 576 $\times$ 4 = 2304} \ 
          $\textbf{(100\%)}$}\\
         \textcolor{gray}{LLaVA-Next-7B} & \textcolor{gray}{64.25}  & \textcolor{gray}{67.9} &  \textcolor{gray}{1852} & \textcolor{gray}{86.5} & \textcolor{gray}{70.35} & \textcolor{gray}{64.84} & \textcolor{gray}{518} &
         \textcolor{gray}{70.17} &
         \textcolor{gray}{58.58} &
         \textcolor{gray}{100.0\%} \\
          \hline

       \rowcolor{mygray}
           \multicolumn{11}{c}{\textit{Retain 11.1\% Tokens} \ $\fg{(\downarrow 88.9\%)}$} \\

        FastV \scriptsize{(ECCV24)} & 57.49 & 54.5 & 1686 &76.9 & \underline{68.22} &49.84 &264 &61.2 &55.35 & 84.02\%   \\
        DART \scriptsize{(EMNLP25)}  & \underline{59.91} &63.5 & \underline{1718} &81.7 &68.12 & \underline{55.65} & \bf357 &63.19 & \underline{57.16} & \underline{90.35\%} \\
        DivPrune \scriptsize{(CVPR25)}& 59.83 & \underline{63.9} & \bf1746 & \underline{83.6} &67.8 &49.32 & \underline{317} & \underline{65.28} &56.46 &89.02\%   \\

        \rowcolor{orange!20}
        Ours & \bf61.38 & \bf64.1 & 1709 & \bf84.3 & \bf68.32 & \bf56.04 & \bf357 & \bf66.2 & \bf57.37 &\bf91.60\%   \\

        \bottomrule[1.5pt]
	\end{tabular}
        \vspace{-2mm}
	\caption{Comparative experiments on image understanding are performed on LLaVA-1.5-13B and LLaVA-Next-7B.}
    \label{tab:2}
\end{table*}

\definecolor{mygray}{gray}{.92}
\definecolor{ForestGreen}{RGB}{34,139,34}
\definecolor{Forestred}{RGB}{220,50,50}
\begin{table*}[!ht]
    \centering
    \setlength{\tabcolsep}{5pt}
    \renewcommand{\arraystretch}{0.9}
    \footnotesize
    \centering
    \begin{tabular}{c | c c c c c c c c c| >{\centering\arraybackslash}p{1.0cm}}
        \toprule[1.5pt]
        \textbf{Method} & \makecell{\textbf{GQA}\\\scriptsize{EM}}& \makecell{\textbf{MMB}\\\scriptsize{Acc.}} & \makecell{\textbf{MME}\\\scriptsize{P+C-score}} & \makecell{\textbf{POPE}\\\scriptsize{F1}} & \makecell{{\textbf{SQA}}\\\scriptsize{IMG-EM}} &  \makecell{{\textbf{VQA}$^{\text{Text}}$}\\\scriptsize{EM}} & \makecell{\textbf{OCRBench}\\\scriptsize{Acc.}} & \makecell{{\textbf{SEED}$^{\text{Image}}$}\\\scriptsize{Acc.}} & \makecell{\textbf{OK-VQA}\\\scriptsize{EM}}& \makecell[c]{\textbf{Avg}.}\\
        \hline
        \rowcolor{mygray}
          \multicolumn{11}{c}{\textit{Fixed Resolution (Min pixels = Max pixels = 1280 $\times$ 28 $\times$ 28), Dynamic Height and Width}}\\
          \rowcolor{mygray}
          \multicolumn{11}{c}{\textit{Upper Bound, About 1350 Tokens} \ 
          $\textbf{(100\%)}$}\\
         \textcolor{gray}{Qwen2-VL-7B-Instruct} & \textcolor{gray}{61.64}  & \textcolor{gray}{81.6} &  \textcolor{gray}{2321} & \textcolor{gray}{88.5} & \textcolor{gray}{85.42} & \textcolor{gray}{82.73} & \textcolor{gray}{796} &
         \textcolor{gray}{76.65} &
         \textcolor{gray}{50.21} &
         \textcolor{gray}{100.0\%} \\
          \hline
       \rowcolor{mygray}
         \multicolumn{11}{c}{\textit{Retain About 450 Tokens} \ $\fg{(\downarrow 66.7\%)}$} \\

        FastV \scriptsize{(ECCV24)} & 60.10 & \bf79.6 & \bf2319 & 87.2 & 82.94 & \bf81.28 & 646 & 73.52 & 49.43 & 96.04\% \\
        DART \scriptsize{(EMNLP25)} & 59.55 & 78.4 & 2259 & 86.9 & 82.25 & 78.06 & 651 & 72.54 & \bf49.83 & 94.95\% \\
        DivPrune \scriptsize{(CVPR25)} & \underline{61.11} & 79.4 &  2199 & \bf88.0 & \underline{83.39}  & 80.07& \underline{704}& \bf75.03 & 49.1 &   \underline{96.57\%} \\

        \rowcolor{orange!20}
        Ours & \bf61.54 & \underline{79.5} & \underline{2273} & \underline{87.9} & \bf84.28 & \underline{81.01}& \bf750 & \underline{74.90} & \underline{49.76} & \bf98.02\%  \\

        \hline
       \rowcolor{mygray}
         \multicolumn{11}{c}{\textit{Retain About 300 Tokens} \ $\fg{(\downarrow 77.8\%)}$} \\

        FastV \scriptsize{(ECCV24)} & 58.90 & 77.6 & \bf2290& 86.4 & \underline{82.35}& \underline{79.43}& 569 & 70.95& 48.77 & 93.39\% \\
        DART \scriptsize{(EMNLP25)} & 57.64 & 77.0 & 2181 & 86.0 & 80.47& 74.10& 586 & 70.20 & \underline{48.83} & 91.69\%  \\
        DivPrune \scriptsize{(CVPR25)} & \underline{60.59}  & \underline{77.8} & 2202 & \underline{87.8}& 82.10 & 77.74 & \underline{607} & \underline{73.61} &  48.65 & \underline{94.11\%} \\

        \rowcolor{orange!20}
        Ours & \bf61.49 & \bf78.1 & \underline{2266} & \bf88.2 & \bf83.49 & \bf79.51 & \bf702 & \bf74.09 & \bf49.03& \bf96.57\% \\

        \hline
       \rowcolor{mygray}
           \multicolumn{11}{c}{\textit{Retain About 150 Tokens} \ $\fg{(\downarrow 88.9\%)}$} \\

        FastV \scriptsize{(ECCV24)} & 55.65 & 73.5 & \bf2174 & 81.6 & 78.53 & \underline{72.86} & 454 & 64.79 & 46.59 & 86.73\%  \\
        DART \scriptsize{(EMNLP25)}  & 55.86 & 73.0 & 2087 & 81.6 & 78.14  & 65.10  & \underline{481} & 65.02 & 46.59 &  85.60\% \\
        DivPrune \scriptsize{(CVPR25)}& \underline{59.17} & \underline{74.8} & 2059 & \underline{86.5} & \underline{80.22} & 71.65 & 472 & \underline{70.65} & \underline{47.29} & \underline{88.93\%}  \\

        \rowcolor{orange!20}
        Ours & \bf60.49 & \bf76.1 & \underline{2158} & \bf87.4 & \bf80.81 & \bf76.15 & \bf581 & \bf72.14 & \bf48.35 & \bf92.58\%  \\

        \bottomrule[1.5pt]
	\end{tabular}
	\caption{Comparative experiments on image understanding are performed on Qwen2-VL-7B-Instruct.}
    \label{tab:3}
\end{table*}

\subsection{Benchmarks and Base VLMs}
To comprehensively evaluate the capability of our proposed VLM-Pruner, which is aimed at multimodal reasoning and understanding, we conduct extensive experiments on a diverse set of benchmark datasets and VLMs with varying architectures. Specifically, we select a broad range of commonly used tasks and benchmarks covering multiple-choice question answering (QA), optical character recognition (OCR), and open-ended QA. The evaluation includes 9 image-language benchmarks, namely GQA \cite{gqa}, MMB \cite{mmbench}, MME \cite{mme}, POPE \cite{pope}, SQA \cite{sqa}, TextVQA \cite{textvqa}, OCRBench \cite{ocrbench}, SEEDBench \cite{seed}, and OKVQA \cite{okvqa}, as well as 4 video-language benchmarks, namely EgoSchema \cite{egoschema}, NExTQA \cite{nextqa}, VideoMME \cite{videomme}, and EgoPlan \cite{egoplan}.
For base VLMs, we choose image-based models including LLaVA-1.5-7B \cite{llava15}, LLaVA-1.5-13B \cite{llava15}, LLaVA-Next-7B \cite{llavanext}, and Qwen2-VL-7B \cite{qwen2vl}, together with the video-based model LLaVA-Video-7B \cite{llavavideo}. Please refer to the supplementary materials for more details.

\subsection{Implementation Details}
For LLaVA-1.5~\cite{llava} with a fixed number of visual tokens, we set $H = W = 24$ ($N = 576$), so that $D_{\max} = 24\sqrt{2}$, which exactly matches the normalization in Eq.~\eqref{eq:bss}. 
For LLaVA-Next~\cite{llavanext} and Qwen2-VL~\cite{qwen2vl}, which adopt a dynamic number of tokens depending on the image resolution, $D_{\max}$ is computed according to the actual height and width of the token feature map. 
For LLaVA-Video~\cite{llavavideo}, which involves three spatial-temporal dimensions (height, width, and frames), the computation is correspondingly performed in the 3D coordinate space. 
Meanwhile, we set $\lambda = 0.5$, $\tau^{(0)} = 0.8$, $\Delta\tau = 0.1$, $q = 256$, $\beta = 0.3$, and $B = 16$. 

All experiments are performed on a setup consisting of 8 GPUs, each with 32 GB of VRAM. 
We adhere to the official evaluation protocols established for both LLaVA-1.5 and LLaVA-Next. Additionally, for evaluating Qwen2-VL and LLaVA-Video, we utilize the \texttt{lmms-evals} package~\cite{lmmseval}.
All results are obtained with a batch size of 1. 

\subsection{Comparative Experiment}
\label{compare}
We select importance-driven baselines, such as FastV \cite{fastv}, SparseVLM \cite{sparsevlm}, PDrop \cite{pyramiddrop}, MustDrop \cite{mustdrop}, and CDPruner \cite{cdpruner}, as well as redundancy-reduction baselines, such as DART \cite{dart}, DivPrune  \cite{divprune}, SAINT \cite{saint}, and BTP \cite{btp}, as the comprehensive baselines for comparison. We also adapt VLM-Pruner to Qwen3-VL-4B and conduct additional comparisons in the Supplementary Material.

\subsubsection{Image-Language Understanding}
\noindent\textbf{LLaVA-1.5-7B.} We evaluate VLM-Pruner under a fixed visual token budget, as shown in Table~\ref{tab:1}. Across 192, 128, and 64 tokens, our pruning preserves \emph{98.85\%}, \emph{98.07\%}, and \emph{95.61\%} of the upper bound Avg. At 64 tokens with $\downarrow88.9\%$ pruning, VLM-Pruner ranks first on seven benchmarks and remains competitive under extreme sparsity. The proximity prior in BSS concentrates selections on informative regions and retains fine grained details, which yields a favorable performance and sparsity tradeoff and increasing gains over baselines as the budget tightens.

\noindent\textbf{LLaVA-1.5-13B.} We study the larger capacity setting with a fixed visual token budget, as shown in Table~\ref{tab:2}. At 64 tokens with $\downarrow88.9\%$ pruning, our approach attains an Avg.\ of \emph{92.68\%} and exceeds DivPrune by \emph{+2.48\%}, DART by \emph{+4.56\%}, and FastV by \emph{+7.93\%}. VLM-Pruner also maintains first place on seven benchmarks, which indicates that redundancy and detail balancing transfers to higher capacity VLMs without loss of relative advantage. The relative margins widen on categories that require fine-grained grounding (GQA) and OCR recognition (OCRBench), suggesting that VLM-Pruner better preserves evidence with more comprehensive details than redundancy-based methods.

\noindent\textbf{LLaVA-Next-7B.} We examine the dynamic resolution setting where the token count varies with image size, as shown in Table~\ref{tab:2}. With $\downarrow88.9\%$ pruning, VLM-Pruner attains an Avg.\ of \emph{91.60\%} and ranks first on eight benchmarks. This demonstrates strong adaptability to fluctuations in token count induced by resolution changes, which is essential for maintaining coverage of local evidence as the visual grid varies, thereby reducing performance loss relative to redundancy-based methods.

\noindent\textbf{Qwen2-VL-7B.} We evaluate architectural generality across three pruning rates, as reported in Table~\ref{tab:3}. The Avg.\ improvement increases with sparsity and reaches \emph{+3.65\%} at $\downarrow88.9\%$. On OCRBench, which is sensitive to small-size text, VLM-Pruner attains a Final score of \emph{581} versus \emph{481} for DART, an absolute gain of \emph{+12.56\%}. These results indicate that sparsity-aware selection preserves fine-grained visual cues and remains effective across architectures.

\definecolor{mygray}{gray}{.92}
\definecolor{ForestGreen}{RGB}{34,139,34}
\definecolor{Forestred}{RGB}{220,50,50}
\begin{table}[t]
    \centering
    \setlength{\tabcolsep}{1pt}
    \renewcommand{\arraystretch}{0.9}
        \footnotesize
        \centering
        \begin{tabular}{c | c c c c  | >{\centering\arraybackslash}p{1.0cm}}
            \toprule[1.5pt]
            \textbf{Method} & \makecell{\textbf{EgoSch.}\\\scriptsize{Acc.}}& \makecell{\textbf{NExTQA}\\\scriptsize{EM/WUPS}} & \makecell{\textbf{VideoMME}\\\scriptsize{P-score}} & \makecell{\textbf{EgoPlan}\\\scriptsize{Acc.}}& \makecell[c]{\textbf{Avg}.}\\
            \rowcolor{mygray}
              \multicolumn{6}{c}{\textit{Max frames = 16}}\\
              \rowcolor{mygray}
              \multicolumn{6}{c}{\textit{Upper Bound, Max Tokens = 16 $\times$ 182 = 2912} \ 
              $\textbf{(100\%)}$}\\
             \textcolor{gray}{LLaVA-Video-7B} & \textcolor{gray}{55.44}  & \textcolor{gray}{78.75/34.11} &  \textcolor{gray}{59.85} & \textcolor{gray}{33.80} & \textcolor{gray}{100\%}\\
              \hline
    
           \rowcolor{mygray}
               \multicolumn{6}{c}{\textit{Retain $Frame \times 20$  Tokens} \ $\fg{(\downarrow 88.9\%)}$} \\
    
            FastV \scriptsize{(ECCV24)} & 43.97  &  67.91/30.66 & 50.44 & 28.27 & 84.96\%\\
            DART \scriptsize{(EMNLP25)}  & \bf48.36 & \underline{72.84}/31.66 & 53.59 & \underline{28.90} & \underline{89.84\%}\\
            DivPrune \scriptsize{(CVPR25)}& 46.27  & 72.42/\underline{31.87} & \underline{53.96} & 27.73 & 88.52\%\\
    
            \rowcolor{orange!20}
            Ours & \underline{47.33} & \bf72.93/32.02 & \bf54.96 & \bf29.58 & \bf90.55\% \\
    
            \bottomrule[1.5pt]
    	\end{tabular}
    	\caption{Comparative experiments on image understanding are performed on LLaVA-Video-7B-Qwen2.}
        \label{tab:4}

\end{table}

\subsubsection{Video-Language Understanding}
\noindent\textbf{LLaVA-Video-7B-Qwen2.} We evaluate the video and language capability with $\downarrow88.9\%$ pruning that retains 20 tokens per frame from the original $13\times14=182$ tokens, as shown in Table~\ref{tab:4}. VLM-Pruner surpasses prior work across four benchmarks and attains NExTQA \emph{72.93/32.02}, giving absolute improvements of \emph{+0.09} EM and \emph{+0.15} WUPS over the second best. On VideoMME it reaches a P-score of \emph{54.96}, an absolute gain of \emph{+1.00}. On EgoPlan it achieves an Accuracy of \emph{29.58\%}, an absolute gain of \emph{+0.68\%}. The overall Avg.\ is \emph{90.55\%} versus \emph{89.84\%} for the second best, an absolute gain of \emph{+0.71\%}, which demonstrates robustness to severe token reduction in spatiotemporal reasoning. BSS computed in 3D coordinates supports stable selection across height, width, and time and avoids dispersed yet incomplete token sets that appear in redundancy-based strategies.

\subsection{Efficiency Analysis}
We analyze the efficiency of VLM-Pruner on LLaVA-1.5-7B and Qwen2-VL-7B. Specifically, we report FLOPs following \cite{fastv} and total inference time on the POPE and OCRBench benchmarks under a $\downarrow$88.9\% pruning rate (i.e., retaining 11.1\% of visual tokens), as shown in Table~\ref{tab:efficiency}. VLM-Pruner outperforms the importance-based baselines SparseVLM and MustDrop in both speedup and FLOPs. In comparisons with redundancy-based methods, DivPrune exhibits slightly fewer FLOPs. This is because it prunes at layer 0, while VLM-Pruner and DART start at layer 2. However, this DivPrune setting comes at the cost of significantly poorer performance compared to our method (Sec. \ref{compare}). VLM-Pruner's inference speed matches DART. On Qwen2-VL-7B, which has more raw visual tokens, VLM-Pruner excels in detail-intensive tasks like OCRBench. It achieves the fastest speed (1.60×) and the best performance (Table \ref{tab:3}). These results demonstrate VLM-Pruner’s superior capability for handling high-resolution inputs.

\begin{table}[t]
    \centering
    \setlength{\tabcolsep}{0.5pt}
    \renewcommand{\arraystretch}{0.95}
    \footnotesize
    \begin{tabular}{@{}lccccc}
        \toprule[1.2pt]
         \multirow{2}{*}{\textbf{Method}} & \multirow{2}{*}{\textbf{Visual Tokens $\downarrow$}} & \multicolumn{2}{c}{\textbf{Total Time $\downarrow$ (Speedup$\uparrow$) }} & \multirow{2}{*}{\textbf{FLOPs $\downarrow$}}    \\
           && \textbf{POPE} & \textbf{OCRBench} & \\
         \midrule
         \textcolor{gray}{LLaVA-1.5-7B} & \textcolor{gray}{576} & \textcolor{gray}{33:21} & \textcolor{gray}{7:14} & \textcolor{gray}{100\%}  \\
         \hspace{0.5em} + SparseVLM & 64 & 24:57 (1.34$\times$) & 9:34 (0.76$\times$) & 23.14\%  \\
         \hspace{0.5em} + MustDrop & 64 & 27:31 (1.21$\times$) & 8:10 (0.89$\times$) & 22.39\%   \\
         \hspace{0.5em} + DART & 64 & 24:06 (1.38$\times$) & 5:57 (1.22$\times$) & 22.09\%   \\
         \hspace{0.5em} + DivPrune & 64 & 22:11 (1.50$\times$) & 5:45 (1.26$\times$) & 16.89\%  \\
          \rowcolor{orange!20}
         \hspace{0.5em} + Ours & 64 & 23:59 (1.39$\times$)  & 6:04 (1.19$\times$) & 22.09\%\\
       \midrule
       \textcolor{gray}{Qwen2-VL-7B} & \textcolor{gray}{100\%} & \textcolor{gray}{11:33:16} & \textcolor{gray}{1:28:16} & \textcolor{gray}{100\%}  \\
         \hspace{0.5em} + DART & 11.1\% & 7:17:20 (1.59$\times$) & 58:16 (1.51$\times$) & 19.51\%   \\
         \hspace{0.5em} + DivPrune & 11.1\% & 6:55:46 (1.67$\times$) & 56:33 (1.56$\times$) & 13.32\%  \\
        \rowcolor{orange!20}
         \hspace{0.5em} + Ours & 11.1\% & 7:21:44 (1.57$\times$)  & 55:07 (1.60$\times$) & 19.51\%\\
        \bottomrule[1.2pt]
    \end{tabular}
    \caption{Inference costs of the number of tokens, Total-Time, Speedup, and FLOPs.}
    \label{tab:efficiency}
\end{table}


\subsection{Ablation Study}
\paragraph{Structural decomposition analysis (Table~\ref{tab:jiegou}).}
\textbf{(i) Stage 1.} Replacing the max-min diverse pivot initialization with a direct Top-4 L1-distance keys seeding the Avg.\ by \emph{-1.27\%}, indicating that semantically diverse pivots cover the target region better than simply selecting large-magnitude key vectors.
\textbf{(ii) Stage 2.} Substituting the selector of stage 2 with DART or DivPrune reduces the Avg.\ to \emph{90.46\%} and \emph{46.47\%}, respectively, showing clear degradation because pure redundancy removal disperses coverage and loses detail. Removing the normalized nearest-distance term in BSS (i.e., omitting $\bar{\delta}_i(S)$) yields a drop of \emph{-1.11\%}, which underscoring the benefit of centrifugal expansion for preserving details.
\textbf{(iii) Stage 3.} Dropping similarity-weighted aggregation lowers the Avg.\ from \emph{95.30\%} to \emph{95.07\%}. Although the gain is modest, it is consistent and improves performance without increasing the token count, so it is complementary to the first two stages.

\begin{figure}[t]
    \vspace{-2mm}
    \centering
    \includegraphics[width=\linewidth]{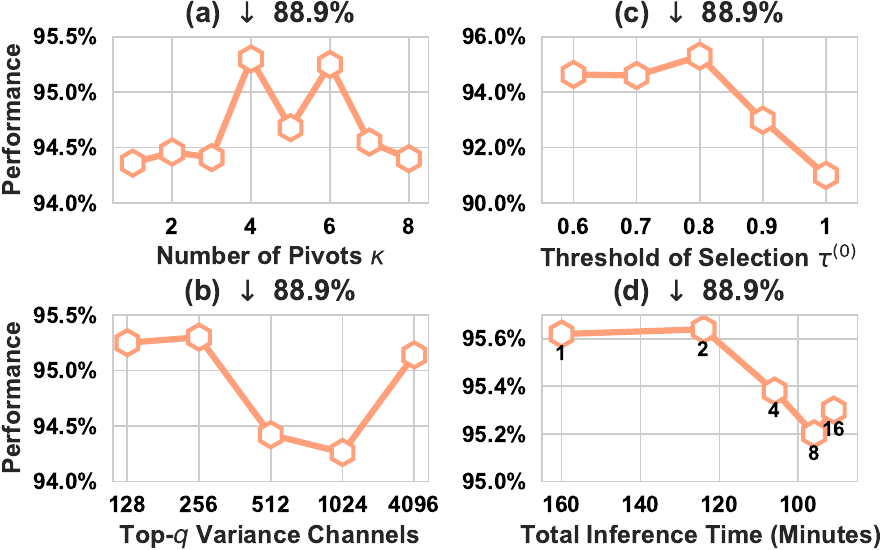}
    \caption{\textbf{Ablation studies on hyperparameters on LLaVA-1.5-7B.} (a) Number of pivots $\kappa$, (b) Top-$q$ highest variance channels, (c) $\tau^{(0)}$, (d) Total inference time (affected by token batch size $B$).}
    \vspace{-5mm}
    \label{fig:ablation}
\end{figure}

\paragraph{Ablation studies on hyperparameters (Fig.~\ref{fig:ablation}).}
\textbf{(a)} Four pivots are most robust. Fewer fail to cover multiple targets. More tend to over-cover edges.
\textbf{(b)} Variance-based channel selection is best around $q=256$. Larger $q$ adds redundancy and complexity.
\textbf{(c)} The initial threshold $\tau^{(0)}=0.8$ performs best. A smaller threshold admits far candidates too early and causes dispersion. A larger threshold increases iterations and hinders effective expansion.
\textbf{(d)} Increasing token batch $B$ reduces latency while performance remains stable. $B=16$ offers a good speed-performance trade-off.

\renewcommand{\multirowsetup}{\centering}
\definecolor{mygray}{gray}{.92}
\definecolor{ForestGreen}{RGB}{34,139,34}
\definecolor{Forestred}{RGB}{220,50,50}
\begin{table}[t]
    \centering
    \setlength{\tabcolsep}{0.2pt}
    \renewcommand{\arraystretch}{0.9}
    \footnotesize
    \centering
    \begin{tabular}{c | c c c c c c| >{\centering\arraybackslash}p{1.0cm}}
        \toprule[1.5pt]
        \textbf{Method} & \textbf{MME} & \textbf{POPE} & \textbf{VQA}$^{\text{T}}$ & \textbf{OCRBench} &  \textbf{SEED}$^{\text{I}}$ & \textbf{OK-VQA} & \makecell[c]{\textbf{Avg}.}\\
        \hline
        \rowcolor{mygray}
        LLaVA-1.5-7B & \multicolumn{7}{c}{\textit{Upper Bound, 576  Tokens} \ $\textbf{(100\%)}$}\\
         \textcolor{gray}{Vanilla} & \textcolor{gray}{1862} & \textcolor{gray}{85.9} & \textcolor{gray}{58.17} & \textcolor{gray}{297} & \textcolor{gray}{66.18} & \textcolor{gray}{57.98} &  \textcolor{gray}{100.0\%} \\
        \rowcolor{orange!20} Ours & 1752 & \textbf{82.2} & \textbf{56.05} & \textbf{279} & \textbf{62.24} & \textbf{56.63}  &  \textbf{95.30\%} \\
          \hline
       \rowcolor{mygray}
        Stage 1 & \multicolumn{7}{c}{\textit{Retain 64 Tokens} \ $\fg{(\downarrow 88.9\%)}$} \\

        Top-4 keys & 1754 & 80.8 & 55.61 & 266 & 61.90 & 56.37  & 94.03\%  \\

        \hline
        \rowcolor{mygray}
        Stage 2 & \multicolumn{7}{c}{\textit{Retain 64 Tokens} \ $\fg{(\downarrow 88.9\%)}$} \\
        + DART & 1727 & 75.8 & 53.15 & 248 & 60.41 & 55.42 & 90.46\%\\
        + DivPrune & 1000 & 28.8 & 42.5 & 24 & 37.50  & 31.17 & 46.47\% \\
        w/o $\bar{\delta}_i(S)$ & 1726 & 81.9 & 55.54 & 270 & 61.82 & 56.41  & 94.19\%  \\
        \hline
        \rowcolor{mygray}
        Stage 3 & \multicolumn{7}{c}{\textit{Retain 64 Tokens} \ $\fg{(\downarrow 88.9\%)}$} \\
        w/o stage 3 & \textbf{1795} & 81.2 & 55.72 & 275 & 61.91 & 56.56  & 95.07\%  \\
        \bottomrule[1.5pt]
	\end{tabular}
	\caption{Structural decomposition analysis in three stages.}
    \label{tab:jiegou}
\end{table}

\section{Conclusion}
We present VLM-Pruner, a training-free Centrifugal token pruning paradigm that explicitly balances redundancy and spatial sparsity to preserve fine-grained visual details while aggressively reducing compute. By combining a BSS criterion with parallel greedy selection and similarity-weighted aggregation, our approach attains strong end-to-end speedups without sacrificing understanding quality. Extensive experiments across 5 VLMs and 13 image/video benchmarks show consistent state-of-the-art performance under severe pruning, with particular robustness on detail-sensitive tasks such as grounding and OCR. We believe these establish an effective foundation for the practical and high-efficiency multimodal reasoning.

{
    \small
    \bibliographystyle{ieeenat_fullname}
    \bibliography{main}
}

\clearpage
\setcounter{page}{1}
\maketitlesupplementary

\section{More Comparative Experiment}
\subsection{More Baselines}
we select CDPruner \cite{cdpruner}, BTP \cite{btp}, and SAINT \cite{saint} as baselines for more theoretical and empirical comparisons with VLMPruner. To further demonstrate VLMPruner's practical timeliness and applicability, we adapt it to the Qwen3-VL-4B vision-language model and conduct additional comparisons. As shown in Table \ref{tab:qwen3}, VLMPruner exhibits a clear advantage while retaining only 11.1\% of visual tokens.
Compared to other methods, \textbf{(1) CDPruner \cite{cdpruner}} focuses on \textbf{importance-driven} pruning by selecting tokens based on joint relevance to both text and visual tokens, but it may miss critical local details. \textbf{(2) BTP \cite{btp}}, while \textbf{ redundancy-reduction}, disrupts fine-grained details early in the process with shallow-layer pruning, rendering later importance-driven pruning less effective. \textbf{(3) SAINT \cite{saint}}, relying purely on \textbf{redundancy-reduction} pruning, may lose important local features due to its lack of spatial consideration. VLMPruner balances diversity and detail completeness by prioritizing local information through centrifugal expansion. This selects more relevant tokens while preserving fine-grained details, resulting in a more balanced selection process that excels in fine-grained visual tasks such as OCR and outperforming the second-best SAINT method by 2.15\%.

\definecolor{mygray}{gray}{.92}
\definecolor{ForestGreen}{RGB}{34,139,34}
\definecolor{Forestred}{RGB}{220,50,50}
\begin{table}[!ht]
    \centering
    \setlength{\tabcolsep}{1pt}
    \renewcommand{\arraystretch}{0.9}
    \footnotesize
    \centering
    \scalebox{0.8}{
    \begin{tabular}{c | c c c c c c c c| >{\centering\arraybackslash}p{1.0cm}}
        \toprule[1.5pt]
        \textbf{Method} & \textbf{GQA}& \textbf{MMB} & \textbf{MME}& \textbf{POPE}& \textbf{SQA} &  \textbf{VQA}$^{\text{Text}}$ & \textbf{OCRBench} & \textbf{OK-VQA}& \makecell[c]{\textbf{Avg}.}\\
        \hline
         \textcolor{gray}{Qwen3-VL-4B} & \textcolor{gray}{62.97 }  & \textcolor{gray}{84.02} &  \textcolor{gray}{2327 } & \textcolor{gray}{90.10 } & \textcolor{gray}{92.56 } & \textcolor{gray}{77.94 } & \textcolor{gray}{835 } &
         \textcolor{gray}{47.42 } &
         \textcolor{gray}{100.0\%} \\
          \hline
       \rowcolor{mygray}
           \multicolumn{10}{c}{\textit{Retain About 120 Tokens} \ $\fg{(\downarrow 88.9\%)}$} \\

        SAINT &  \underline{58.64} & \textbf{81.11}  & 2025  &  83.13 &  \textbf{85.87} &  \underline{63.75} & \underline{566}  &\underline{41.59}   &  \underline{87.38\%}  \\
        BTP &  42.03  & 56.22  &  1424 & 61.00  & 76.95  &  55.45 &  431 &  13.27 &  62.06\% \\
        CDPruner &  57.24  & 78.08  &  \underline{2042} & \underline{86.59}  & 86.12  &  63.39 & 486  & 35.18  &  84.31\%  \\

        \rowcolor{orange!20}
        Ours & \textbf{59.72}  & \underline{78.48}  & \textbf{2129}  & \textbf{88.51}  &  \underline{85.23} & \textbf{68.37}  & \textbf{573} &  \textbf{42.60} &  \textbf{89.53\%}  \\

        \bottomrule[1.5pt]
	\end{tabular}
    }
	\caption{Comparative experiments on image understanding are performed on Qwen3-VL-4B-Instruct.}
    \label{tab:qwen3}
\end{table}

\subsection{More Visualizations}
\begin{figure}[h]
    \centering
    \includegraphics[width=\linewidth]{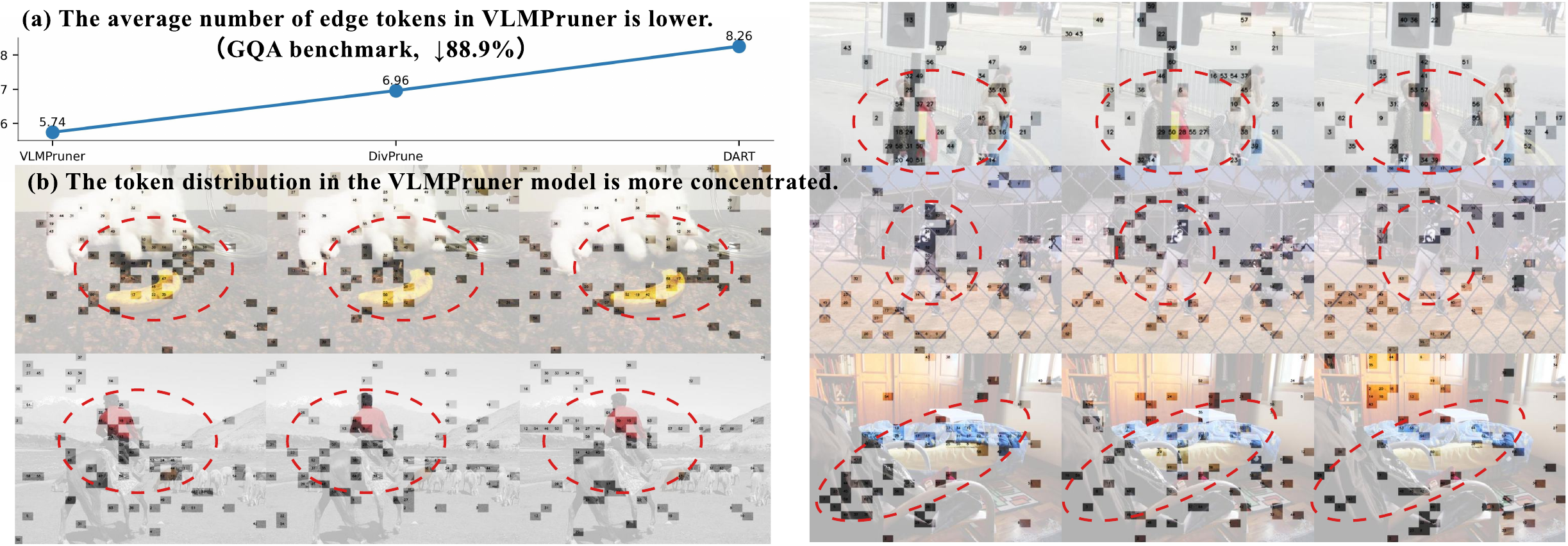}
    \caption{More visualizations of the actual pruning effects between baselines and VLM-Pruner. From left to right are VLM-Pruner, DivPrune, and DART. (a) The average number of edge tokens in VLM-Pruner is lower. (b) The token distribution in the VLM-Pruner model is more concentrated.}
    \vspace{-2mm}
    \label{fig:supp1}
\end{figure}

VLMPruner’s centrifugal expansion design is based on the assumption that visual tokens exhibit spatial locality, prioritizing the retention of important information within local neighborhoods. By balancing diversity and detail completeness through the combination of redundancy and spatial sparsity, VLMPruner ensures that nearby tokens are more likely to be selected, avoiding detail loss from dispersed coverage. As shown in Fig. \ref{fig:supp1}, this reduces edge tokens' number and improves detail completeness, leading significant improvements in tasks like fine-grained recognition. Firstly, because edge tokens exhibit extremely low similarity, DART and DivPrune select them far more frequently than VLMPruner. Moreover, VLMPruner preserves fine-grained details more effectively than DART and DivPrune, such as regions highlighted by the red circles in Fig. \ref{fig:supp1}.

\section{More Ablation Study}
To further validate the effectiveness of token pruning by VLM-Pruner and the reasonableness of the hyperparameter settings, we present several additional experimental results on LLaVA-1.5-7B with the pruning rate of 88.9\%, focusing on (a) number of pivots $\kappa$, (b) top-$q$ highest variance channels, (c) threshold $\tau^{(0)}$ of token selection, (d) token batch size $B$, (e) aggregation weight $\beta$, and (f) pruning layer $i$.

\subsection{Ablation Study on the Number of Pivots $\kappa$}
In the experiment on the number of pivots as shown in Table \ref{tab:sm1}, we observe that the performance varied with different values of $\kappa$. The optimal performance was achieved with $\kappa = 4$, resulting in an average performance of 95.30\%. This suggests that four pivots strike the best balance, providing sufficient coverage of the target regions without overfitting to a specific area. Fewer pivots (e.g., $\kappa = 1$ or $\kappa = 2$) fail to capture enough semantic diversity, while too many pivots (e.g., $\kappa = 7$ or $\kappa = 8$) tend to overfit and capture redundant tokens from less important regions.

\renewcommand{\multirowsetup}{\centering}
\definecolor{mygray}{gray}{.92}
\definecolor{ForestGreen}{RGB}{34,139,34}
\definecolor{Forestred}{RGB}{220,50,50}
\begin{table}[h]
    \centering
    \setlength{\tabcolsep}{0.2pt}
    \renewcommand{\arraystretch}{0.9}
    \footnotesize
    \centering
    \begin{tabular}{c | c c c c c c| >{\centering\arraybackslash}p{1.0cm}}
        \toprule[1.5pt]
        \textbf{Method} & \textbf{MME} & \textbf{POPE} & \textbf{VQA}$^{\text{T}}$ & \textbf{OCRBench} &  \textbf{SEED}$^{\text{I}}$ & \textbf{OK-VQA} & \makecell[c]{\textbf{Avg}.}\\
        \hline
        \rowcolor{mygray}
        LLaVA-1.5-7B & \multicolumn{7}{c}{\textit{Upper Bound, 576  Tokens} \ $\textbf{(100\%)}$}\\
         \textcolor{gray}{Vanilla} & \textcolor{gray}{1862} & \textcolor{gray}{85.9} & \textcolor{gray}{58.17} & \textcolor{gray}{297} & \textcolor{gray}{66.18} & \textcolor{gray}{57.98} &  \textcolor{gray}{100.0\%} \\
          \hline
       \rowcolor{mygray}
        $\kappa$ & \multicolumn{7}{c}{\textit{Retain 64 Tokens} \ $\fg{(\downarrow 88.9\%)}$} \\
        1 & \underline{1737} & 81.6  & 55.93  & 271 & 61.83  & 56.29   &  94.36\%  \\
        2 & 1729 & 81.9 &  55.66 &  271 & \underline{62.33} &  56.52 &  94.46\%  \\
        3 & 1725 & \bf82.4  &  55.80 & 269 & \bf62.36  &  56.33  &  94.41\%  \\
        \rowcolor{orange!20}
        4 & \bf1752 & \underline{82.2}  &  \underline{56.05} & \underline{279} & 62.24  &  56.63  &  \bf95.30\%  \\
        5 & 1734 & 81.6  &  55.97 & 273 & 62.21  &  \underline{56.72}  &  94.68\%  \\
        6 & 1732 & 82.0  &  \bf56.11 & \bf281 & 62.17  &  \bf56.81  &  \underline{95.25\%}  \\
        7 & 1722 & 81.5  &  55.88 & 274 &  62.11 &  56.67  &  94.55\%  \\
        8 & 1707 & 82.0  &  56.03 & 273 &  62.13 &  56.33  &  94.40\%  \\
        \bottomrule[1.5pt]
	\end{tabular}
	\caption{Ablation study on the number of pivots $\kappa$.}
    \label{tab:sm1}
\end{table}

\subsection{Ablation Study on the Top-$q$ Highest Variance Channels}
For the experiment on selecting the top-$q$ highest variance channels as shown in Table \ref{tab:sm2}, we find that setting $q=256$ achieved the best performance, with an average score of 95.30\%. This configuration retains enough channels to preserve essential visual information while minimizing redundancy. Larger values of $q$ result in diminishing returns, as they introduce more redundant information and computational complexity, which doesn't contribute significantly to performance improvement. On the other hand, smaller values of $q$ lead to the loss of critical visual features, reducing the overall model performance. Therefore, $q=256$ is the optimal setting, offering a balanced trade-off between computational efficiency and model performance.

\renewcommand{\multirowsetup}{\centering}
\definecolor{mygray}{gray}{.92}
\definecolor{ForestGreen}{RGB}{34,139,34}
\definecolor{Forestred}{RGB}{220,50,50}
\begin{table}[h]
    \centering
    \setlength{\tabcolsep}{0.2pt}
    \renewcommand{\arraystretch}{0.9}
    \footnotesize
    \centering
    \begin{tabular}{c | c c c c c c| >{\centering\arraybackslash}p{1.0cm}}
        \toprule[1.5pt]
        \textbf{Method} & \textbf{MME} & \textbf{POPE} & \textbf{VQA}$^{\text{T}}$ & \textbf{OCRBench} &  \textbf{SEED}$^{\text{I}}$ & \textbf{OK-VQA} & \makecell[c]{\textbf{Avg}.}\\
        \hline
        \rowcolor{mygray}
        LLaVA-1.5-7B & \multicolumn{7}{c}{\textit{Upper Bound, 576  Tokens} \ $\textbf{(100\%)}$}\\
         \textcolor{gray}{Vanilla} & \textcolor{gray}{1862} & \textcolor{gray}{85.9} & \textcolor{gray}{58.17} & \textcolor{gray}{297} & \textcolor{gray}{66.18} & \textcolor{gray}{57.98} &  \textcolor{gray}{100.0\%} \\
          \hline
       \rowcolor{mygray}
        Top-$q$ & \multicolumn{7}{c}{\textit{Retain 64 Tokens} \ $\fg{(\downarrow 88.9\%)}$} \\
        128 & \bf1765  & 81.8   &  \bf56.25  & \underline{277}  &  62.02 & \underline{56.70}  &   \underline{95.25\%}  \\
        \rowcolor{orange!20}
        256 & \underline{1752}  & \bf82.2   &  \underline{56.05}  &  \bf279 &  \underline{62.24} &  56.63 &   \bf95.30\%  \\
        512 & 1738  &  81.3  &  55.87  & 270  & 62.21  &  56.59 &   94.42\%  \\
        1024 & 1743  &  81.1  &  55.51  & 268  & \bf62.26  & \bf56.71  &  94.26\%  \\
        4096 (raw) & \underline{1752}  &  \underline{81.9}  &  55.97  & \bf279  & 62.18 & 56.41  &   95.14\%  \\
        \bottomrule[1.5pt]
	\end{tabular}
	\caption{Ablation study on the Top-$q$ highest variance channels.}
    \label{tab:sm2}
\end{table}

\subsection{Ablation Study on the Threshold $\tau^{(0)}$ of Token Selection}
In the experiment on the token selection threshold $\tau^{(0)}$ as shown in Table \ref{tab:sm3}, we observe that setting $\tau^{(0)} = 0.8$ resulted in the best overall performance of 95.30\%. This threshold ensures that the model prioritizes selecting tokens that are spatially close to previously selected ones, effectively avoiding the early inclusion of distant or irrelevant tokens. If the threshold is set too low, distant tokens are selected too early, causing the token set to become scattered and less effective in representing fine-grained details. Conversely, setting the threshold too high causes an overemphasis on spatial distance, resulting in a more compact token selection that introduces increased redundancy. Therefore, a threshold of 0.8 strikes a balance between selecting non-redundant tokens and preserving finer-grained details.

\renewcommand{\multirowsetup}{\centering}
\definecolor{mygray}{gray}{.92}
\definecolor{ForestGreen}{RGB}{34,139,34}
\definecolor{Forestred}{RGB}{220,50,50}
\begin{table}[h]
    \centering
    \setlength{\tabcolsep}{0.2pt}
    \renewcommand{\arraystretch}{0.9}
    \footnotesize
    \centering
    \begin{tabular}{c | c c c c c c| >{\centering\arraybackslash}p{1.0cm}}
        \toprule[1.5pt]
        \textbf{Method} & \textbf{MME} & \textbf{POPE} & \textbf{VQA}$^{\text{T}}$ & \textbf{OCRBench} &  \textbf{SEED}$^{\text{I}}$ & \textbf{OK-VQA} & \makecell[c]{\textbf{Avg}.}\\
        \hline
        \rowcolor{mygray}
        LLaVA-1.5-7B & \multicolumn{7}{c}{\textit{Upper Bound, 576  Tokens} \ $\textbf{(100\%)}$}\\
         \textcolor{gray}{Vanilla} & \textcolor{gray}{1862} & \textcolor{gray}{85.9} & \textcolor{gray}{58.17} & \textcolor{gray}{297} & \textcolor{gray}{66.18} & \textcolor{gray}{57.98} &  \textcolor{gray}{100.0\%} \\
          \hline
       \rowcolor{mygray}
        $\tau^{(0)}$ & \multicolumn{7}{c}{\textit{Retain 64 Tokens} \ $\fg{(\downarrow 88.9\%)}$} \\
        0.6 &  \bf1753 &  81.9  &  55.00  &  \underline{276} &  61.82 &  56.48  &  \underline{94.64\%} \\
        0.7 & 1704  &  \bf82.5  &  \underline{55.68}  &  \underline{276} & \underline{61.93}  &  \bf56.71  &  94.60\% \\
        \rowcolor{orange!20}
        0.8 &  \underline{1752} &  \underline{82.2}  &  \bf56.05  &  \bf279 &  \bf62.24 &  \underline{56.63}  &  \bf95.30\% \\
        0.9 &  1719 &  78.3  &  54.71  &  269 & 61.29  &  56.39  &  92.99\% \\
        1.0 & 1725  &  73.9  &  53.66  & 265  & 59.87 & 55.31   &  91.00\% \\
        \bottomrule[1.5pt]
	\end{tabular}
	\caption{Ablation study on the threshold $\tau^{(0)}$ of token selection.}
    \label{tab:sm3}
\end{table}

\subsection{Ablation Study on the Token Batch Size $B$}
In the experiment on token batch size $B$ as shown in Table \ref{tab:sm4}, the results show that a batch size of $B=16$ provides the best performance and fastest inference speed, with an average score of 95.30\% and a total time of 91 minutes. Larger batch sizes, such as $B=32$, may reduce inference time, but they may lead to the failure of the BSS criterion due to fewer iterations. Smaller batch sizes result in longer inference times, even slower than the initial LLaVA-1.5-7B at 131 minutes. The token batch size of $B=16$ offers an optimal trade-off, minimizing latency while maintaining the quality of the selected tokens.

\renewcommand{\multirowsetup}{\centering}
\definecolor{mygray}{gray}{.92}
\definecolor{ForestGreen}{RGB}{34,139,34}
\definecolor{Forestred}{RGB}{220,50,50}
\begin{table}[h]
    \centering
    \setlength{\tabcolsep}{0.2pt}
    \renewcommand{\arraystretch}{0.9}
    \footnotesize
    \centering
    \begin{tabular}{c c | c c c c c c| >{\centering\arraybackslash}p{1.0cm}}
        \toprule[1.5pt]
        \multicolumn{2}{c|}{\textbf{Method}} & \textbf{MME} & \textbf{POPE} & \textbf{VQA}$^{\text{T}}$ & \textbf{OCRBench} &  \textbf{SEED}$^{\text{I}}$ & \textbf{OK-VQA} & \makecell[c]{\textbf{Avg}.}\\
        \hline
        \rowcolor{mygray}
        \multicolumn{2}{c|}{LLaVA-1.5-7B} & \multicolumn{7}{c}{\textit{Upper Bound, 576  Tokens} \ $\textbf{(100\%)}$}\\
         \textcolor{gray}{Vanilla} & \textcolor{gray}{131} & \textcolor{gray}{1862} & \textcolor{gray}{85.9} & \textcolor{gray}{58.17} & \textcolor{gray}{297} & \textcolor{gray}{66.18} & \textcolor{gray}{57.98} &  \textcolor{gray}{100.0\%} \\
          \hline
       \rowcolor{mygray}
        $B$ & \makecell{Total \\ Time \\ (Min.)} & \multicolumn{7}{c}{\textit{Retain 64 Tokens} \ $\fg{(\downarrow 88.9\%)}$} \\
        1 & 160 & \bf1755 &  \bf83.6  & 56.06  &  275  & \bf62.95   &  \underline{56.88}  &  \underline{95.62\%} \\
        2 & 124 & 1742 &  \underline{83.5}  &  55.96 &  \bf279  &  \underline{62.73}  &  \bf56.93  &  \bf95.64\% \\
        4 & 106 & 1730 &  83.3  & \underline{56.11}  &  \underline{278}  & 62.57   &  56.70  &  95.38\% \\
        8 & \underline{96} & 1726 &  \underline{83.5}  & \bf56.26  &  274  &  62.42  &  56.82  &  95.20\% \\
        \rowcolor{orange!20}
        16 & \bf91 & \underline{1752} & 82.2 &  56.05 &  \bf279  &  62.24  &  56.63  &  95.30\% \\
        \bottomrule[1.5pt]
	\end{tabular}
	\caption{Ablation study on the token batch size $B$.}
    \label{tab:sm4}
\end{table}

\subsection{Ablation Study on the Aggregation Weight $\beta$}

In the aggregation weight $\beta$ experiment, as shown in Table \ref{tab:sm5}, we focus on minimizing the influence of Stage 3 to emphasize the importance of Stage 1 and 2, which manage buffering for spatial sparsity and token selection, while ensuring the preservation of fine-grained details. The results show that $\beta=0.3$ yields an average score of 95.30\%, providing a moderate improvement over the 95.07\% achieved without Stage 3. It also enables effective aggregation of discarded token information without overemphasizing it. This setting further preserves fine-grained details, as demonstrated by the best score of 279 achieved on OCRBench.

\renewcommand{\multirowsetup}{\centering}
\definecolor{mygray}{gray}{.92}
\definecolor{ForestGreen}{RGB}{34,139,34}
\definecolor{Forestred}{RGB}{220,50,50}
\begin{table}[h]
    \centering
    \setlength{\tabcolsep}{0.2pt}
    \renewcommand{\arraystretch}{0.9}
    \footnotesize
    \centering
    \begin{tabular}{c | c c c c c c| >{\centering\arraybackslash}p{1.0cm}}
        \toprule[1.5pt]
        \textbf{Method} & \textbf{MME} & \textbf{POPE} & \textbf{VQA}$^{\text{T}}$ & \textbf{OCRBench} &  \textbf{SEED}$^{\text{I}}$ & \textbf{OK-VQA} & \makecell[c]{\textbf{Avg}.}\\
        \hline
        \rowcolor{mygray}
       LLaVA-1.5-7B & \multicolumn{7}{c}{\textit{Upper Bound, 576  Tokens} \ $\textbf{(100\%)}$}\\
         \textcolor{gray}{Vanilla} & \textcolor{gray}{1862} & \textcolor{gray}{85.9} & \textcolor{gray}{58.17} & \textcolor{gray}{297} & \textcolor{gray}{66.18} & \textcolor{gray}{57.98} &  \textcolor{gray}{100.0\%} \\
          \hline
       \rowcolor{mygray}
        $\beta$ & \multicolumn{7}{c}{\textit{Retain 64 Tokens} \ $\fg{(\downarrow 88.9\%)}$} \\
        0.2 & 1737  &  82.1 &  55.86 & \underline{278} & 62.26 &  56.41 &   94.98\% \\
        \rowcolor{orange!20}
        0.3 & 1752 &  \underline{82.2} &  56.05 & \bf279 & 62.24 &  56.63 &   95.30\% \\
        0.4 & 1754 &  \bf82.4 &  \underline{56.26} & 277 & 62.28 & 56.72  &   95.34\% \\
        0.5 & 1773 &  \bf82.4 &  \bf56.44 & \underline{278} & \bf62.36 & 56.80  &   \underline{95.66\%} \\
        0.6 & 1790 &  82.0 &  56.16 & \underline{278} & \bf62.36 & \bf56.87  &   \bf95.68\% \\
        0.7 & \underline{1799} &  81.5 & 56.23 & 277 & \underline{62.33} & 56.80  &   95.60\% \\
        0.8 & \bf1808 &  81.5 & 56.07 & 275 & 62.14 & \underline{56.81}  &   95.47\% \\
        \hline
        w/o stage 3 & 1795 & 81.2 & 55.72 & 275 & 61.91 & 56.56  & 95.07\%  \\
        \bottomrule[1.5pt]
	\end{tabular}
	\caption{Ablation study on the aggregation weight $\beta$.}
    \label{tab:sm5}
\end{table}

To further emphasize the core innovation of VLM-Pruner, the BSS criterion, we validate the model performance on Qwen-VL-7B after removing Stage 3, as shown in Table \ref{tab:sm7}. The results show that even without Stage 3, our method still outperforms existing baselines by an absolute margin of approximately 5\%. This highlights the effectiveness of the proposed centrifugal token pruning paradigm.

\renewcommand{\multirowsetup}{\centering}
\definecolor{mygray}{gray}{.92}
\definecolor{ForestGreen}{RGB}{34,139,34}
\definecolor{Forestred}{RGB}{220,50,50}
\begin{table}[h]
    \centering
    \setlength{\tabcolsep}{0.2pt}
    \renewcommand{\arraystretch}{0.9}
    \footnotesize
    \centering
    \begin{tabular}{c | c c c c c c| >{\centering\arraybackslash}p{1.0cm}}
        \toprule[1.5pt]
        \textbf{Method} & \textbf{MME} & \textbf{POPE} & \textbf{VQA}$^{\text{T}}$ & \textbf{OCRBench} &  \textbf{SEED}$^{\text{I}}$ & \textbf{OK-VQA} & \makecell[c]{\textbf{Avg}.}\\
        \hline
        \rowcolor{mygray}
       Qwen2-VL-7B & \multicolumn{7}{c}{\textit{Upper Bound} \ $\textbf{(100\%)}$}\\
         \textcolor{gray}{Vanilla} & \textcolor{gray}{2321} & \textcolor{gray}{88.55} & \textcolor{gray}{82.73} & \textcolor{gray}{796} & \textcolor{gray}{76.65} & \textcolor{gray}{50.21} &  \textcolor{gray}{100.0\%} \\
          \hline
       \rowcolor{mygray}
        Stage 3 & \multicolumn{7}{c}{\textit{Retain 11.1\% Tokens} \ $\fg{(\downarrow 88.9\%)}$} \\
        w/o Stage 3 &  \bf2197 &   86.4  &  \underline{74.89}  & \underline{561} & 70.69 &  \underline{48.20} &   \underline{90.25\%} \\
         \hline
         FastV & \underline{2174} & 81.6 & 72.86 & 454 & 64.79 & 46.59 & 84.72\% \\
         DART & 2087 & 81.6 & 65.10 & 481 & 65.02 & 46.59 & 83.14\% \\
         DivPrune & 2059 & \underline{86.5} & 71.65 & 472 & \underline{70.75} & 47.29 & 86.47\% \\
        \rowcolor{orange!20}
        Ours &  2158 & \bf87.4  &  \bf76.15 & \bf581 & \bf72.14 & \bf48.35  &   \bf91.20\% \\
        \bottomrule[1.5pt]
	\end{tabular}
	\caption{Additional structural decomposition analysis.}
    \label{tab:sm7}
\end{table}

\subsection{Ablation Study on the Pruning Layer $i$}
In the layer-wise pruning experiment summarized in Table \ref{tab:sm6}, pruning at layer 2 achieves the best balance and an average performance of 95.30\% by successfully preserving fine-grained visual details while reducing task-irrelevant redundancy. Unlike DivPrune, VLM-Pruner cannot prune layer 0 because it relies on token $keys$ from the previous layer to perform coarse semantic abstraction. Pruning at layer 1 leads to notable performance degradation, likely because the text and visual tokens have not yet fully interacted, which hampers the model's ability to focus on prompt-relevant regions and results in suboptimal token selection. Conversely, pruning at later layers such as layer 3 or 4 may remove more less important details, and this also comes at the cost of lower model performance and increased computational overhead in FLOPs.

\renewcommand{\multirowsetup}{\centering}
\definecolor{mygray}{gray}{.92}
\definecolor{ForestGreen}{RGB}{34,139,34}
\definecolor{Forestred}{RGB}{220,50,50}
\begin{table}[h]
    \centering
    \setlength{\tabcolsep}{0.2pt}
    \renewcommand{\arraystretch}{0.9}
    \footnotesize
    \centering
    \begin{tabular}{c | c c c c c c| >{\centering\arraybackslash}p{1.0cm}}
        \toprule[1.5pt]
        \textbf{Method} & \textbf{MME} & \textbf{POPE} & \textbf{VQA}$^{\text{T}}$ & \textbf{OCRBench} &  \textbf{SEED}$^{\text{I}}$ & \textbf{OK-VQA} & \makecell[c]{\textbf{Avg}.}\\
        \hline
        \rowcolor{mygray}
       LLaVA-1.5-7B & \multicolumn{7}{c}{\textit{Upper Bound, 576  Tokens} \ $\textbf{(100\%)}$}\\
         \textcolor{gray}{Vanilla} & \textcolor{gray}{1862} & \textcolor{gray}{85.9} & \textcolor{gray}{58.17} & \textcolor{gray}{297} & \textcolor{gray}{66.18} & \textcolor{gray}{57.98} &  \textcolor{gray}{100.0\%} \\
          \hline
       \rowcolor{mygray}
        layer $i$ & \multicolumn{7}{c}{\textit{Retain 64 Tokens} \ $\fg{(\downarrow 88.9\%)}$} \\
        1 & 1689  &  82.0   &  55.68  & 267 & 61.54 &  56.30 &   93.65\% \\
        \rowcolor{orange!20}
        2 & \underline{1752}  &  82.2   &  \bf56.05  & \bf279 & \underline{62.24} &  \bf56.63 &   \bf95.30\% \\
        3 &  1725 &  \underline{82.9}  &  \underline{55.85}  & \underline{274} & \bf62.39 &  \underline{56.53} &   94.86\% \\
        4 & \bf1759  &  \bf83.3  &  55.70  & 270 & 62.18 &  56.48 &   \underline{94.91\%} \\
        \bottomrule[1.5pt]
	\end{tabular}
	\caption{Ablation study on the pruning layer $i$.}
    \label{tab:sm6}
\end{table}

The choice of the pruning layer is based on the following reasons: (1) According to the Align-KD \cite{alignkd}, feature similarity between layers after the second layer exceeds 0.9, meaning the second layer already contains rich contextual information. Stage 3 compensates for unselected token information, minimizing information loss. (2) Pruning is performed on just one layer, resulting in low computational cost and time consumption. (3) It is easy and fast to adapt to different VLMs with default parameters. In contrast, methods like PDrop \cite{pyramiddrop}, BTP \cite{btp}, and SAINT \cite{saint} require manual tuning of pruning layers and thresholds, making adaptation more complex.

\subsection{Ablation Study on other base VLMs}
To further validate the "plug-and-play" claim of our method, we conduct additional ablation experiments on Qwen3-VL-4B. We select the most critical stages, Stage 1 and Stage 2, for partial benchmarking experiments. As shown in Table \ref{tab:qwen3_ab}, the results confirm the strong advantages of the default parameters presented in the paper.

\definecolor{mygray}{gray}{.92}
\definecolor{ForestGreen}{RGB}{34,139,34}
\definecolor{Forestred}{RGB}{220,50,50}
\begin{table}[!ht]
    \centering
    \setlength{\tabcolsep}{1pt}
    \renewcommand{\arraystretch}{0.9}
    \footnotesize
    \centering
    \scalebox{0.8}{
    \begin{tabular}{c | c |c c |c c c| c c  c | c >{\centering\arraybackslash}p{1.0cm}}
        \toprule[1.5pt]
        \textbf{Benchmark} & \textbf{Qwen3-VL-4B} & \textbf{Query}& \textbf{Value}& \textbf{$\tau=0.6$} & \textbf{$0.7$} & \textbf{$0.9$}&\textbf{$\lambda=0.3$} & \textbf{$0.7$} & \textbf{$0.9$} & \textbf{Ours}\\
        \hline
       \rowcolor{mygray}
           \multicolumn{11}{c}{\textit{Retain About 120 Tokens} \ $\fg{(\downarrow 88.9\%)}$} \\

        MME & 2327  &  2065 & 2055  & 2089 & 2096 & 2106 &  2074 & 2095  &  2096 &  \textbf{2129} \\
        SQA &  92.56  & 84.53  &  85.12 & 84.23 & 84.58 & 84.53  & 85.18  & 84.18  &  84.73 & \textbf{85.23} \\
        OCRBench &  835  &  571 & 551 & 502 &  550 & 544  & 559  & 557  &  552 &   \textbf{573} \\

        \bottomrule[1.5pt]
	\end{tabular}
    }
	\caption{Ablation Study on image understanding are performed on Qwen3-VL-4B-Instruct.}
    \label{tab:qwen3_ab}
\end{table}

\section{Details for Benchmarks}
We select a broad range of commonly used tasks and benchmarks. Specifically, the evaluation includes 9 image-language benchmarks, namely GQA \cite{gqa}, MMB \cite{mmbench}, MME \cite{mme}, POPE \cite{pope}, SQA \cite{sqa}, TextVQA \cite{textvqa}, OCRBench \cite{ocrbench}, SEEDBench \cite{seed}, and OKVQA \cite{okvqa}, as well as 4 video-language benchmarks, namely EgoSchema \cite{egoschema}, NExTQA \cite{nextqa}, VideoMME \cite{videomme}, and EgoPlan \cite{egoplan}.

\subsection{Image-Language Benchmarks}

\paragraph{GQA} GQA is designed to evaluate visual scene understanding and compositional reasoning capabilities of VLMs on real-world imagery.

\paragraph{MMB} MMBench provides a hierarchical evaluation framework for VLMs, organizing 20 distinct capability dimensions into three progressive levels (L1 to L3).

\paragraph{MME} MME assesses both perceptual and cognitive abilities of VLMs, where perception includes optical character recognition (OCR), coarse- and fine-grained object recognition, and cognition encompasses commonsense reasoning, numerical calculation, text translation, and code reasoning.

\paragraph{POPE} POPE quantifies object hallucination in VLMs by reformulating the problem as a binary judgment task on the presence of objects in an image.

\paragraph{SQA} ScienceQA measures scientific reasoning proficiency across natural science, social science, and language science domains, supplemented with detailed explanations and lectures.

\paragraph{TextVQA} TextVQA evaluates the integration of visual and textual information, requiring models to read and reason about text present in images.

\paragraph{OCRBench} OCRBench provides a comprehensive evaluation of text-related capabilities, covering representative tasks such as scene text recognition, document-oriented understanding, key information extraction, and handwritten mathematical expression recognition.

\paragraph{SEEDBench} SEEDBench assesses multimodal understanding across text and visual modalities, including scene classification, object detection, and attribute recognition.

\paragraph{OKVQA} OKVQA examines the ability of VLMs to leverage external knowledge, as the image content alone is insufficient to answer the questions.

\subsection{Video-Language Benchmarks}

\paragraph{EgoSchema} EgoSchema focuses on long-term video-language understanding, requiring capabilities such as scene comprehension, object state tracking, and extended visual memory.

\paragraph{NExTQA} NExTQA challenges VLMs to interpret temporal and causal structures in videos, emphasizing reasoning about action causality, temporal progression, and object interactions.

\paragraph{VideoMME} VideoMME is the first benchmark dedicated to holistic video analysis, systematically evaluating model performance across varying video durations (short, medium, long) and input modalities (video frames, subtitles, audio).

\paragraph{EgoPlan} EgoPlan assesses VLMs as embodied task planners, involving interactions with hundreds of objects under complex visual settings to evaluate the prediction of feasible actions.

\end{document}